\documentclass[10pt,letterpaper]{article}
\usepackage[top=0.85in,left=2.75in,footskip=0.75in]{geometry}

\usepackage{amsmath,amssymb}
\usepackage{makecell}

\usepackage{changepage}
\usepackage{textcomp}

\usepackage[utf8x]{inputenc}

\usepackage{textcomp,marvosym}

\usepackage{cite}
\usepackage{nameref,hyperref}
\usepackage{float}
\usepackage{multirow}
\usepackage{arydshln}
\usepackage[right]{lineno}

\usepackage{microtype}
\DisableLigatures[f]{encoding = *, family = * }

\usepackage{array}
\usepackage{lscape}
\usepackage{longtable}
\usepackage{rotating}
\usepackage{url,algorithm,algpseudocode}
\newcolumntype{+}{!{\vrule width 2pt}}

\newlength\savedwidth

\raggedright
\usepackage[aboveskip=1pt,labelfont=bf,labelsep=period,justification=raggedright,singlelinecheck=off]{caption}

\makeatletter
\renewcommand{\@biblabel}[1]{\quad#1.}
\makeatother

\usepackage{lastpage,fancyhdr,graphicx}
\usepackage{epstopdf}
\fancyheadoffset[L]{2.25in}
\fancyfootoffset[L]{2.25in}
\usepackage[color=orange, textsize=small]{todonotes}
\reversemarginpar
\usepackage{schemata}
\newcommand\AB[2]{\schema{\schemabox{#1}}{\schemabox{#2}}}

\usepackage{tikz}
\usetikzlibrary{shapes, arrows, arrows.meta}
\usetikzlibrary{trees}

\begin{document}
\vspace*{0.2in}

\begin{flushleft}
{\Large
\textbf\newline{A reproducible experimental survey on biomedical sentence similarity: a string-based method sets the state of the art} 
}
\newline
\\
Alicia Lara-Clares\textsuperscript{1*},
Juan J. Lastra-D\'{i}az\textsuperscript{1},
Ana Garcia-Serrano\textsuperscript{1}
\\
\bigskip
\textbf{1} NLP \& IR Research Group, E.T.S.I. Inform\'{a}tica, Universidad Nacional de Educaci\'{o}n a Distancia (UNED), Madrid (Spain)
\\
\bigskip

* alara@lsi.uned.es

\end{flushleft}
\section*{Abstract}

This registered report introduces the largest, and for the first time, reproducible experimental survey on biomedical sentence similarity with the following aims: (1) to elucidate the state of the art of the problem; (2) to solve some reproducibility problems preventing the evaluation of most of current methods; (3) to evaluate several unexplored sentence similarity methods; (4) to evaluate for the first time an unexplored benchmark, called Corpus-Transcriptional-Regulation (CTR); (5) to carry out a study on the impact of the pre-processing stages and Named Entity Recognition (NER) tools on the performance of the sentence similarity methods; and finally, (6) to bridge the lack of software and data reproducibility resources for methods and experiments in this line of research. Our reproducible experimental survey is based on a single software platform, which is provided with a detailed reproducibility protocol and dataset as supplementary material to allow the exact replication of all our experiments and results. In addition, we introduce a new aggregated string-based sentence similarity method, called LiBlock, together with eight variants of current ontology-based methods, and a new pre-trained word embedding model trained on the full-text articles in the PMC-BioC corpus. Our experiments show that our novel string-based measure sets the new state of the art on the sentence similarity task in the biomedical domain and significantly outperforms all the methods evaluated herein, with the only exception of one ontology-based method. Likewise, our experiments confirm that the pre-processing stages, and the choice of the NER tool for ontology-based methods, have a very significant impact on the performance of the sentence similarity methods. We also detail some drawbacks and limitations of current methods, and warn on the need of refining the current benchmarks. Finally, a noticeable finding is that our new string-based method significantly outperforms all state-of-the-art Machine Learning (ML) models evaluated herein.


\section*{Introduction}

Measuring semantic similarity between sentences is an important task in the fields of Natural Language Processing (NLP), Information Retrieval (IR), and biomedical text mining, among others. For instance, the estimation of the degree of semantic similarity between sentences is used in text classification \cite{Tafti2017-lq,Kim2012-ex,Chen2017-zj}, question answering \cite{Sarrouti2017-eu,Kosorus2012-bg}, evidence sentence retrieval to extract biological expression language statements \cite{Ravikumar2017-hv,Rastegar-Mojarad2016-ez}, biomedical document labeling \cite{Du2019-rt}, biomedical event extraction \cite{Liu2013-jo}, named entity recognition \cite{Hahn2020-os}, evidence-based medicine \cite{Kim2011-vp,Hassanzadeh2015-hc}, biomedical document clustering \cite{Boyack2011-iq}, prediction of adverse drug reactions \cite{Dey2018-fi}, entity linking \cite{Lamurias2019-ht}, document summarization \cite{Aliguliyev2009-by,Shang2011-zv} and sentence-driven search of biomedical literature \cite{Allot2019-yr}, among other applications. In the question answering task, Sarrouti and El Alaomi \cite{Sarrouti2017-eu} build a ranking of plausible answers by computing the similarity scores between each biomedical question and the candidate sentences extracted from a knowledge corpus. Allot et al. \cite{Allot2019-yr} introduce a system to retrieve the most similar sentences in the BioC biomedical corpus \cite{Comeau2019-vd} called Litsense \cite{Allot2019-yr}, which is based on the comparison of the user query with all sentences in the aforementioned corpus. Likewise, the relevance of the research in this area is endorsed by the proposal of recent conference series, such as SemEval \cite{Agirre2012-gu,Agirre2013-ae,Agirre2014-mf,Agirre2015-gt,Agirre2016-nn,Cer2017-zi} and BioCreative/OHNLP \cite{Wang2018-iw}, and works based on sentence similarity measures, such as the work of Aliguliyev \cite{Aliguliyev2009-by} in automatic document summarization, which shows that the performance of these applications depends significantly on the sentence similarity measures used. 

The aim of any semantic similarity method is to estimate the degree of similarity between two textual semantic units as perceived by a human being, such as words, phrases, sentences, short texts, or documents. Unlike sentences from the language in general use whose vocabulary and syntax is limited both in extension and complexity, most sentences in the biomedical domain are comprised of a huge specialized vocabulary made up of all sort of biological and clinical terms, in addition to an uncountable list of acronyms, which are combined in complex lexical and syntactic forms.

Nowadays, there exist several works in the literature that experimentally evaluate multiple methods on biomedical sentence similarity. However, they are either theoretical or have a limited scope and cannot be reproduced. For instance, Kalyan et al. \cite{Kalyan2020-yj}, Khattak et al. \cite{Khattak2019-rs}, and Alsentzer et al. \cite{Alsentzer2019-hj} introduce theoretical surveys on biomedical embeddings with a limited scope. On the other hand, the experimental surveys introduced by Sogancioglu et al. \cite{Sogancioglu2017-rc}, Blagec et al. \cite{Blagec2019-nl}, Peng et al. \cite{Peng2019-cc}, and Chen et al. \cite{Chen2018-uh} among other authors, cannot be reproduced because of the lack of source code and data to replicate both methods and experiments, or the lack of a detailed definition of their experimental setups. Likewise, there are other recent works whose results need to be confirmed. For instance, Tawfik and Spruit \cite{Tawfik2020-uo} experimentally evaluate a set of pre-trained language models, whilst Chen et al. \cite{Chen2020-mp} propose a system to study the impact of a set of similarity measures on a Deep Learning ensembled model, which is based on a Random Forest model \cite{Breiman2001-mi}. 

The main aim of this work is to introduce a comprehensive and very detailed reproducible experimental survey of methods on biomedical sentence similarity to elucidate the state of the problem by implementing our previous registered report protocol \cite{Lara-Clares2021-av}.  Our experiments are based on our software implementation and evaluation of all methods analyzed herein into a common and new software platform based on an extension of the Half-Edge Semantic Measures Library (HESML) \cite{Lastra-Diaz2017-qo, lastra-diaz2022}, called HESML\footnote{\url{http://hesml.lsi.uned.es}}  for Semantic Textual Similarity (HESML-STS). All our experiments have been recorded into a Docker virtualization image that is provided as supplementary material together with our software \cite{EPNXTR_2021Docker} and a detailed reproducibility protocol \cite{Lara-Clares2022protocolsIO} and dataset \cite{EPNXTR_2021Dataset} to allow the easy replication of all our methods, experiments, and results. This work is based on our previous experience developing reproducible research in a series of publications in the area, such as the experimental surveys on word similarity introduced in \cite{Lastra-Diaz2015-xc, Lastra-Diaz2015-ct, Lastra-Diaz2016-or, Lastra-Diaz2019-ai}, whose reproducibility protocols and datasets \cite{Lastra-Diaz2016-ag, Lastra-Diaz2019-oj} are detailed and independently confirmed in two companion reproducible papers \cite{Lastra-Diaz2017-qo, Lastra-Diaz2021-kl}, and a reproducible benchmark on semantic measures libraries for the biomedical domain \cite{lastra-diaz2022}. Finally, we refer the reader to our previous work \cite{Lara-Clares2021-av} for a very detailed review of the literature on sentence similarity measures, which is omitted herein because of the lack of room and to avoid being redundant.

\subsection*{Main motivations and research questions}

Our main motivation is the lack of a comprehensive and reproducible experimental survey on biomedical sentence similarity that allows setting the state of the problem in a sound and reproducible way, as detailed in our previous registered report protocol \cite{Lara-Clares2021-av}.  Our main research questions are as follows:

\begin{description}
\item[RQ1] Which methods get the best results on biomedical sentence similarity?
\item[RQ2] Is there a statistically significant difference between the best-performing methods and the remaining ones?
\item[RQ3] What is the impact of the biomedical Named Entity Recognition (NER) tools on the performance of the methods on biomedical sentence similarity?
\item[RQ4] What is the impact of the pre-processing stage on the performance of the methods on biomedical sentence similarity?
\item[RQ5] What are the main drawbacks and limitations of current methods on biomedical sentence similarity?
\end{description}

A second motivation is implementing a set of unexplored methods based on adaptations from other methods proposed for the general language domain. A third motivation is the evaluation in the same software platform of the three known benchmarks on biomedical sentence similarity reported in the literature as follows: the Biomedical Semantic Similarity Estimation System (BIOSSES) \cite{Sogancioglu2017-rc} and  Medical Semantic Textual Similarity (MedSTS) \cite{Wang2018-oj} datasets, as well as the evaluation for the first time of the Microbial Transcriptional Regulation (CTR) \cite{Lithgow-Serrano2019-si} dataset in a sentence similarity task, despite it having been previously evaluated in other related tasks, such as the curation of gene expressions from scientific publications \cite{Lithgow-Serrano2020-bx}. A fourth motivation is a study on the impact of the pre-processing stage and NER tools on the performance of the sentence similarity methods, such as that done by Gerlach et al. \cite{Gerlach2019-ay} for stop-words in topic modeling task. And finally, our fifth motivation is the lack of reproducibility software and data resources on this task, which allow an easy replication and confirmation of previous methods, experiments, and results in this line of research, as well as encouraging the development and evaluation of new sentence similarity methods.

\subsection*{Definition of the problem and contributions}

The two main research problems tackled in this work are the design and implementation of a large and reproducible experimental survey on sentence similarity measures for the biomedical domain, and the evaluation of a set of unexplored methods based on adaptations from previous methods used in the general language domain. Our main contributions are as follows: (1) the largest, and for the first time, reproducible experimental survey on biomedical sentence similarity; (2) the first collection of self-contained and reproducible benchmarks on biomedical sentence similarity; (3) the evaluation of a set of previously unexplored methods, such as a new string-based sentence similarity method, based on Li et al. \cite{Li2006-au} and Block distance \cite{Krause1986-wb}, eight variants of the current ontology-based methods from the literature based on the work of Sogancioglu et al. \cite{Sogancioglu2017-rc}, and a new pre-trained Word Embedding (WE) model based on FastText \cite{Bojanowski2017-pb} and trained on the full-text of articles in the PMC-BioC corpus \cite{Comeau2019-vd}; (4) the evaluation for the first time of an unexplored benchmark, called CTR \cite{Lithgow-Serrano2019-si}; (5) the study on the impact of the pre-processing stage and Named Entity Recognition (NER) tools on the performance of the sentence similarity methods; (6) the integration for the first time of most sentence similarity methods for the biomedical domain into the same software library, called HESML-STS, which is available both in Github \footnote{\url{https://github.com/jjlastra/HESML/tree/HESML-STS_master_dev}} and in a reproducible dataset \cite{EPNXTR_2021Dataset}; (7) a detailed reproducibility protocol together with a collection of software tools and datasets provided as supplementary material to allow the exact replication of all our experiments and results; and finally, (8) an analysis of the drawbacks and limitations of the current state-of-the-art methods. 

The rest of the paper is structured as follows. First, we introduce a collection of new sentence similarity methods evaluated herein for the first time. Next, we describe a detailed experimental setup for our experiments on biomedical sentence similarity and introduce our experimental results. Then, we discuss our results and answer the research questions detailed above. Subsequently, we introduce our conclusions and future work. Finally, we introduce three appendices with supplementary material as follows. Appendix A introduces all statistical significance results of our experiments, whilst Appendix B introduces all data tables reporting the performance of all methods with all pre-processing configurations evaluated herein, and the Appendix C introduces a reproducibility protocol detailing a set of step-by-step instructions to allow the exact replication of all our experiments, which is published at \url{protocols.io} \cite{Lara-Clares2022protocolsIO}. 

\section*{The new sentence similarity methods}

This section introduces a new string-based sentence similarity method based on the aggregation of the Li et al. \cite{Li2006-au} similarity and Block distance \cite{Krause1986-wb} measures, called LiBlock, as well as eight new variants of the ontology-based methods proposed by Sogancioglu et al. \cite{Sogancioglu2017-rc}, and a new pre-trained word embedding model based on FastText \cite{Bojanowski2017-pb} and trained on the full-text of the articles in the PMC-BioC corpus \cite{Comeau2019-vd}. 

\subsection*{The new LiBlock string-based method}

Two key advantages of the family of string-based methods are as follows. Firstly, they can be very efficiently computed because they do not require the use of external knowledge or pre-trained models, and secondly, they obtain competitive results as shown in table \ref{tab:table_results}. However, the string-based methods do not capture the semantics of the words in the sentence, which prevent them from recognizing semantic relationships between words, such as synonymy and meronymy among others. On the other hand, the family of ontology-based methods capture the semantic relationships between words in a sentence pair and obtain state-of-the-art results in the sentence similarity task for the biomedical domain, as shown in table \ref{tab:table_results}. However, the effectiveness of ontology-based methods depends on the lexical coverage of the ontologies and the ability to recognize automatically the underlying concepts in sentences by using Named Entity Recognition (NER) and Word Sense Desambiguation (WSD) tools, whose coverage and performance could be limited in several application domains. Precisely, the NER task is still an open problem \cite{Song2021-cc} in the biomedical domain because of the vast biomedical vocabulary and the complex lexical and syntactic forms found in the biomedical literature. Otherwise, the methods based on pre-trained word embedding models provide a broader lexical coverage than the ontology-based ones and obtain better results. However, the methods based on word embeddings do not significantly outperform all ontology-based measures in a word similarity task \cite{Lastra-Diaz2019-ai} in addition to requiring large corpus for training, a complex training phase, and more computational resources than the families of string-based and ontology-based methods.

To overcome the drawbacks and limitations of the string-based and ontology-based methods detailed above, we propose here a new aggregated string-based measure called LiBlock and denoted by $sim_{LiBk}$ henceforth, which is based on the combination of a similarity measure derived from the Block Distance \cite{Krause1986-wb} and an adaptation from the ontology-based similarity measure introduced by Li et al. \cite{Li2006-au} that removes the use of ontologies, such as WordNet \cite{Miller1995-am} or Systematized Nomenclature of Medicine Clinical Terms (SNOMED-CT) \cite{Donnelly2006-ye}. The LiBlock similarity measure obtains the best results in combination with the cTAKES NER tool \cite{Savova2010-ed}, which allows the detection of synonyms of CUI concepts. Nevertheless, the LiBlock method obtains competitive results regarding the state-of-the-art methods with no use, either implicitly or explicitly, of an ontology, as detailed in table \ref{tab:table_results_liblock}.

The $sim_{LiBk}$ method detailed in equation (\ref{eq:LiBlock}) is defined by the linear aggregation of an adaptation of the Li et al. \cite{Li2006-au} measure, called $sim_{LiAd}$ (\ref{eq:LiAd}), and a similarity measure derived from the Block Distance measure \cite{Krause1986-wb}, called $sim_{Bk}$ (\ref{eq:BlockDist}). Let be $L_\Sigma$ the set of word sequences in a universal unseen alphabet $\Sigma$, the $sim_{LiBk}$ function returns a value between 0 and 1 which indicates the similarity score between two input sentences, as defined in equation \ref{eq:LiBlock}. The  $sim_{Bk}$ function is based on the computation of the word frequencies $fr(w_i, s_j)$ for each input sentence $s_1$ and $s_2$ and their concatenation $s_1 + s_2$, as detailed in equation (\ref{eq:BlockDist}). The auxiliary function $fr(w_i, s_j)$ returns the frequency of a word $w_i$ in the word sequence $s_j$, whilst the function $fr(w_i, s_1 + s_2)$ returns the number of occurrences of the word $w_i$ in the concatenation of the two word sequences, denoted by $s_1 + s_2$. On the other hand, the $sim_{LiAd}$ function takes two word sets obtained by invoking the $\sigma$ function (\ref{eq:wordSet}) with the sentences $s_1$ and $s_2$, and then it computes the cosine similarity of the two binary semantic vectors corresponding to invoke the $\varphi(S_1)$ function (\ref{eq:binVectorConstructor}) with the $\sigma(s_1)$ and $\sigma(s_2)$ word sets. Finally, the $sim_{LiBk}$ score is defined by either the linear combination of $sim_{Bk}$ and $sim_{LiAd}$, as detailed in equation (\ref{eq:LiBlock}), or $sim_{Bk}$ if $sim_{LiAd}$ is 0.

\begin{figure}[h!]
\caption{This figure details the workflow for computing the new LiBlock measure and an example illustrating a use case of the workflow following the steps defined in algorithm \ref{algorithmLiBlock}.}
\centering
\tikzstyle {process} = [rectangle, rounded corners, draw=black, fill=gray!10,
                        minimum width=0.75cm, minimum height=0.75cm,
                        text centered, font=\sffamily]
\tikzstyle {sim} = [rectangle, rounded corners, draw=black, fill=blue!10,
                        minimum width=0.75cm, minimum height=0.75cm,
                        text centered, font=\sffamily]
\tikzstyle {outputdata} = [fill=red!10, trapezium, trapezium left                             angle=80,trapezium right angle=-80, draw=black,
                        minimum height=0.57cm,
                        text centered, font=\sffamily]
\tikzstyle {init} = [circle, draw=black, text centered, font=\sffamily, fill=red!15]
\tikzstyle {decision} = [diamond, draw=black, text centered, font=\sffamily, fill=blue!10]
\tikzstyle {initSents} = [circle, draw=black, fill=green!10, text centered, font=\sffamily, trapezium, trapezium left angle=80, trapezium right angle=-80,minimum width=0.75cm, minimum height=0.75cm]
\tikzstyle{line} = [draw, -latex', font=\sffamily, fill=gray!15]

\begin{tikzpicture}[every node/.style={fill=white, font=\sffamily}, align=center]

\pgfdeclarelayer{bg}    
\pgfsetlayers{bg,main}  

\node (s1) [initSents] at (1,-1) {\normalsize{s1}};
\node (s2) [initSents] at (0,0) {\normalsize{s2}};
\node (S1) [process] at (4,-2) {\normalsize{$S_1$}};
\node (D) [process] at (4,-3.5) {\normalsize{$D = S_1 \cup S_2$}};
\node (S2) [process] at (4,-5) {\normalsize{$S_2$}};
\node (simbk) [sim] at (8,-1) {\normalsize{$sim_{Bk}$}};
\node (b1) [process] at (8,-2) {\normalsize{$b_1$}};
\node (simliad) [sim] at (8,-3.5) {\normalsize{$sim_{LiAd}$}};
\node (b2) [process] at (8,-5) {\normalsize{$b_2$}};
\node (simlibk) [sim] at (11,-3.5) {\normalsize{$sim_{LiBk}$}};
\node (score) [outputdata] at (11,-5) {\normalsize{$sim_{LiBk} score$}};

\node (sigmas1) [fill=white!10] at (2,-2) {$\sigma$};
\node (sigmas2) [fill=white!10] at (2,-5) {$\sigma$};

\node (varphis1) [fill=white!10] at (6,-2) {$\varphi$};
\node (varphis2) [fill=white!10] at (6,-5) {$\varphi$};

\begin{pgfonlayer}{bg}    
\draw[->] (s1) -- (1,-2) -- (S1);
\draw[->] (S1) -- (D);
\draw[->] (S2) -- (D);
\draw[->] (s2) -- (0,-5) -- (S2);
\draw[->] (s1) -- (simbk);
\draw[->] (s2) -- (8,0) -- (simbk);
\draw[->] (S1) -- (b1);
\draw[->] (S2) -- (b2);
\draw[->, dashed] (D) -- (simliad);
\draw[->] (b1) -- (simliad);
\draw[->] (b2) -- (simliad);
\draw[->] (simbk) -- (11,-1) -- (simlibk);
\draw[->] (simliad) -- (simlibk);
\draw[->] (simlibk) -- (score);
\end{pgfonlayer}
\end{tikzpicture}

\begin{align}
\text{Input} &:  \text{Raw } s_1 \leftarrow \text{``Lung tumour formation in mice by oncogenic KRAS requires} \nonumber \\
\quad   & \quad\quad\quad\quad\quad   \text{formation Craf, but not Braf."} \nonumber \\
\quad  & \quad\text{Raw } s_2 \leftarrow \text{``The oncogenic activity of mutant Kras appears dependent"} \nonumber \\
\quad   & \quad\quad\quad\quad\quad  \text{on functional Craf but not on Braf."} \nonumber \\
\text{step 1: }& s_1 \leftarrow \left\{ \text{c0280089, formation, mice, oncogenic, c1537502,} \right.  \nonumber \\  
& \quad \quad \quad \left. \text{requires, formation, craf, c0812241} \right\}  \nonumber \\
& s_2 \leftarrow \left\{ \text{oncogenic, activity, mutant, c1537502, appears,} \right. \nonumber \\   
& \quad \quad \quad \left. \text{dependent, functional, craf, c0812241} \right\}  \nonumber \\
\text{step 2: }& S_1 \leftarrow \left\{ \text{c0280089, formation, mice, oncogenic, c1537502,} \right.  \nonumber \\  
& \quad \quad \quad \left. \text{requires, craf, c0812241} \right\}  \nonumber \\
\text{step 3: }& S_2 \leftarrow \left\{ \text{oncogenic, activity, mutant, c1537502, appears,} \right. \nonumber \\   
& \quad \quad \quad \left. \text{dependent, functional, craf, c0812241} \right\}  \nonumber \\
\text{step 4: }& D \leftarrow \left\{ \text{c0280089, formation, mice, oncogenic, c1537502, requires,} \right.  \nonumber \\  
& \quad\quad\quad \left. \text{craf, c0812241, activity, mutant, appears, dependent, functional} \right\} \nonumber \\
\text{step 5: }& b_1 \leftarrow  \{\text{1, 1, 1, 1, 1, 1, 1, 1, 0, 0, 0, 0, 0}\} \nonumber \\
\text{step 6: }& b_2 \leftarrow  \{\text{0, 0, 0, 1, 1, 0, 1, 1, 1, 1, 1, 1, 1}\} \nonumber \\
\text{step 7: }& sim_{LiAd} \leftarrow  \text{0.471} \nonumber \\
\text{step 8: }& sim_{Bk} \leftarrow  \text{0.444} \nonumber \\
\text{step 9: }& sim_{LiBk} \leftarrow  \text{0.458} \nonumber 
\end{align}
\label{fig:LiBlockPipeline}
\end{figure}
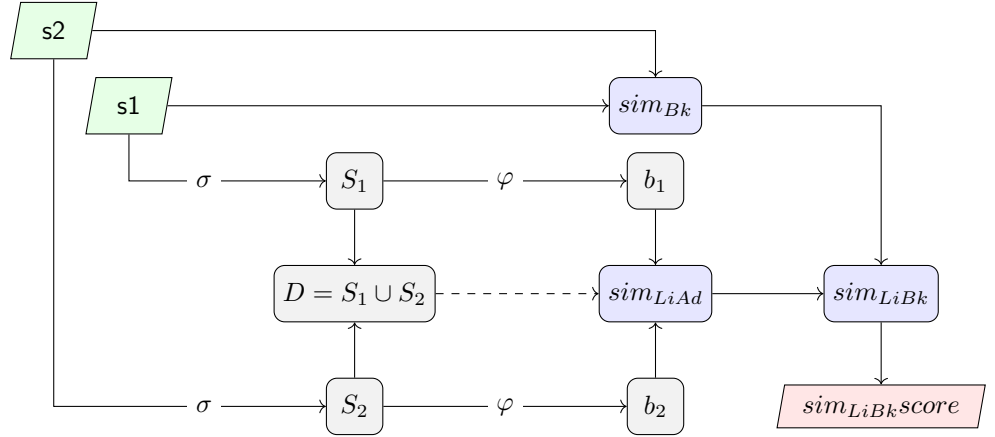

\paragraph{A walk-through example.} Algorithm \ref{algorithmLiBlock} details the step-by-step procedure to compute the $sim_{LiBk}$ function, whilst figure \ref{fig:LiBlockPipeline} shows the pipeline for calculating the LiBlock similarity score defined in equation \ref{eq:LiBlock}, as well as an example for illustrating an end-to-end calculation of the $sim_{LiBk}$ similarity score of two sentences.

\begin{algorithm}[h!]
\begin{algorithmic}[1]
\Function{simLiBlock}{$s_1, s_2$} \Comment{being $s_1, s_2$ word sequences $\in L_\Sigma$}
\State $S_1 \gets \sigma(s_1)$ \Comment{word set sentence 1}
\State $S_2 \gets \sigma(s_2)$ \Comment{word set sentence 2}
\State $D \gets S_1 \cup S_2$ \Comment{construct the dictionary $D$}
\State $b_1 \gets \varphi(S_1)$ \Comment{construct the semantic binary vector $b_1$} 
\State $b_2 \gets \varphi(S_2)$ \Comment{construct the semantic binary vector $b_2$} 
\State $score_{LiAd} \gets sim_{LiAd}(b_1, b_2)$ \Comment{compute LiAdapted similarity}
\State $score_{Bk} \gets sim_{Bk}(s_1, s_2)$ \Comment{compute Block Distance similarity}
\State $score_{LiBk} \gets sim_{LiBk}(score_{LiAd}, score_{Bk})$ \Comment{compute LiBlock similarity}
\State \Return $score_{LiBk}$
\EndFunction
\end{algorithmic}
\caption{LiBlock sentence similarity measure for two input pre-processed sentences.}
\label{algorithmLiBlock}
\end{algorithm}

\begin{align}
& \quad\quad\quad\quad\quad\quad\quad\quad\quad\quad\quad\quad\quad\quad\quad\quad\quad\quad \text{(LiBlock similarity)} \label{eq:LiBlock} \\
sim_{LiBk} &: L_\Sigma\times L_\Sigma \rightarrow [0,1] \subset \mathbb{R}, \text{  } L_\Sigma = \{\text{word sequences in alphabet } \Sigma \} \nonumber \\
\nonumber \\
sim_{LiBk}(s_1, s_2) &= 
\Bigg\{\begin{array}{lr}
sim_{Bk}(s_1,s_2),\quad\quad\quad\quad\quad\quad\quad\quad\quad\text{ if } sim_{LiAd}(\sigma(s_1),\sigma(s_2))=0 \\
\\
\frac{1}{2}sim_{Bk}(s_1,s_2) + \frac{1}{2}sim_{LiAd}(\sigma(s_1), \sigma(s_2)),\quad\quad\quad\text{otherwise}  \\ 
\end{array} \nonumber \\
\nonumber \\
sim_{Bk} &: L_\Sigma \times L_\Sigma \rightarrow [0,1] \subset \mathbb{R}, \quad\quad\quad\quad\quad\quad\quad\quad\quad \text{(Block distance)} \label{eq:BlockDist} \\
sim_{Bk}(s_1,s_2) &=  1 - \frac{\sum\limits_{i=1}^{|D|} |fr(w_i, s_1) - fr(w_i, s_2)|}{\sum\limits_{i=1}^{|D|}fr(w_i, s_1 + s_2)}, \quad D = \sigma(s_1) \cup \sigma(s_2) \in \mathcal{P}(\Sigma)  \nonumber  \\
\nonumber \\
sim_{LiAd} &: \mathcal{P}(D) \times \mathcal{P}(D)  \rightarrow [0,1] \subset \mathbb{R}, \quad\quad\quad\quad\quad\text{(Li's score adaptation)} \label{eq:LiAd} \\
sim_{LiAd}(S_1, S_2) &= \frac{\varphi(S_1) \cdot \varphi(S_2)}{||\varphi(S_1)|| * ||\varphi(S_2)||}   \nonumber \\
\nonumber \\
\varphi &: \mathcal{P}(D) \rightarrow \{0,1\}^{|D|}, \quad\quad\quad\quad\quad\quad\quad\text{(binary vector constructor)}  \label{eq:binVectorConstructor}\\
\varphi(S) &= (b_1, b_2,\dots, b_{|D|}),\quad b_i = \Bigg\{\begin{array}{lr}
1, w_i \in D \\
0, w_i \not\in D 
\end{array} \nonumber \\
\nonumber \\
\sigma &: L_\Sigma \rightarrow \mathcal{P}(\Sigma), \quad\quad\quad\quad\quad\quad\quad\quad\quad\quad\quad\quad\quad\text{(word set generator)} \label{eq:wordSet}\\
\sigma(s) &= \{w \in \Sigma : \exists k \in [1,\text{length}(s)] \text{ such that } s_k = w \}  \nonumber 
\end{align}

\subsection*{The eight new variants of current ontology-based methods}

The current family of ontology-based methods for biomedical sentence similarity proposed by Sogancioglu et al. \cite{Sogancioglu2017-rc} is based on the ontology-based semantic similarity between word and concepts within the sentences to be compared. Thus, this later family of methods defines a framework in which we can design new variants by exploring other word similarity measures. For this reason, we propose herein the evaluation of a set of new ontology-based sentence similarity measures based on two different unexplored notions as follows: (1) the evaluation of state-of-the-art word similarity measures from the general domain \cite{Lastra-Diaz2019-ai} not evaluated in the biomedical domain yet; and (2) the evaluation of several ontology-based word similarity measures based on a recent and very efficient shortest-path algorithm, called Ancestors-based Shortest-Path Length (AncSPL) \cite{lastra-diaz2022}, which is a fast approximation of the Dijkstra's algorithm \cite{Dijkstra1959-ng} for taxonomies that is introduced with the first HESML version for the biomedical domain \cite{lastra-diaz2022}.

Thus, we propose here the evaluation based on the combination of WBSM and UBSM methods with the path-based word similarity methods as follows: WBSM-Rada (M7); WBSM-cosJ\&C (M9); WBSM-coswJ\&C (M10); WBSM-Cai (M11); UBSM-Rada (M12); UBSM-cosJ\&C (M14); UBSM-coswJ\&C (M15); and UBSM-Cai (M16). The detailed information about this later method is shown in table \ref{tab:descriptionmethods_ontology}.

\subsection*{The new pre-trained word embedding model}

Current sentence similarity methods based on the evaluation of pre-trained embedding models are mostly trained using PubMed Central (PMC) Open Access dataset \footnote{\url{https://www.ncbi.nlm.nih.gov/labs/pmc/}}, or Medical Information Mart for Intensive Care (MIMIC-III) clinical notes \cite{Johnson2016-hn}. However, as far as we know, there are no models in the literature trained on the full text of the articles in the PMC-BioC corpus \cite{Comeau2019-vd}. Therefore, we propose evaluating a new FastText \cite{Bojanowski2017-pb} word embedding model trained on the aforementioned BioC corpus. FastText overcomes one significant limitation of other methods, such as word2vec \cite{Mikolov2013-fi} and GloVe \cite{Pennington2014-fk}, which ignore the morphology of words by assigning a vector to each word in the vocabulary. For a more detailed review of the family of word embedding methods, we refer the authors to the recent reproducible survey by Lastra-Díaz et al. \cite{Lastra-Diaz2019-ai}. The configuration parameters for training this model are detailed in table \ref{tab:descriptionmethods_sentenceEmbeddings}, and all the necessary information and resources for evaluating it are available in our reproducibility dataset \cite{EPNXTR_2021Dataset}, as detailed in table \ref{tab:table_material}.

\section*{The reproducible experimental survey}
\label{sec:evaluation}

This section introduces a detailed experimental setup to evaluate and compare all the sentence similarity methods for the biomedical domain proposed in our primary work \cite{Lara-Clares2021-av}, together with the new methods introduced herein. The main aims of our experiments are as follows: (1) the evaluation of most of known methods for biomedical sentence similarity onto the three biomedical datasets shown in table \ref{tab:table_Datasets}, and implemented in the same software platform; (2) the evaluation of a set of new sentence similarity methods adapted from their definitions for the general-language domain; (3) the evaluation of a new sentence method called LiBlock introduced in this work, eight variants of the current ontology-based methods from the literature based on the work of Sogancioglu et al. \cite{Sogancioglu2017-rc}, and a new word embedding model based on FastText and trained on the full-text of articles in the PMC-BioC corpus \cite{Comeau2019-vd}; (4) the setting of the state of the art of the problem in a sound and reproducible way; (5) the replication and independent confirmation of previously reported methods and results; (6) a study on the impact of different pre-processing configurations on the performance of the sentence similarity methods; (7) a study on the impact of different Name Entity Recognition (NER) tools, such as MetaMap \cite{Aronson2010-pb} and clinic Text Analysis and Knowledge Extraction System (cTAKES) \cite{Savova2010-ed}, onto the performance of the sentence similarity methods; (8) the evaluation for the first time of the CTR \cite{Lithgow-Serrano2019-si} dataset; (9) the identification of the main drawbacks and limitations of current methods; and finally, (10) a detailed statistical significance analysis of the results.

\begin{table}[!h]
\caption{Benchmarks on biomedical sentence similarity evaluated in this work.}
\begin{tabular}{lll}
\small{Dataset} & \small{\#pairs} & \small{Corresponding file (*.tsv) in HESML-STS distribution} \\
\hline
\small{BIOSSES \cite{Sogancioglu2017-rc}} & \small{100} & \small{BIOSSESNormalized.tsv} \\
\small{MedSTS \cite{Wang2018-oj}} & \small{1,068} & \small{CTRNormalized\_averagedScore.tsv} \\
\small{CTR \cite{Lithgow-Serrano2019-si}} & \small{170} & \small{MedStsFullNormalized.tsv} \\
\end{tabular}
\label{tab:table_Datasets}
\end{table}

\subsubsection*{Selection of methods} 

The criteria for the selection of the sentence similarity methods evaluated herein is as follows: (a) all the methods that have been evaluated in BIOSSES and MedSTS datasets; (b) a selection of methods that have not been evaluated in the biomedical domain yet; (c) a collection of new variants or adaptations of methods previously proposed for the general or biomedical domain, which are evaluated for the first time in this work, such as the WBSM-cosJ\&C \cite{Sogancioglu2017-rc,Lastra-Diaz2015-ct,Sanchez2011-cf,lastra-diaz2022}, WBSM-coswJ\&C \cite{Sogancioglu2017-rc,Lastra-Diaz2015-ct,Sanchez2011-cf,lastra-diaz2022}, WBSM-Cai \cite{Sogancioglu2017-rc,Cai2017-di,lastra-diaz2022}, UBSM-cosJ\&C \cite{Sogancioglu2017-rc,Lastra-Diaz2015-ct,Sanchez2011-cf,lastra-diaz2022}, UBSM-coswJ\&C \cite{Sogancioglu2017-rc,Lastra-Diaz2015-ct,Sanchez2011-cf,lastra-diaz2022}, and UBSM-Cai \cite{Sogancioglu2017-rc,Cai2017-di,lastra-diaz2022} methods detailed in tables \ref{tab:descriptionmethods_ontology} and  \ref{tab:descriptionmethods_sentenceEmbeddings}; and (d) a new string-based method based on Li et al. \cite{Li2006-au} introduced in this work. For a more detailed description of the selection criteria of the methods, we refer the reader to our registered report protocol \cite{Lara-Clares2021-av}.

Tables \ref{tab:descriptionmethods_string} and ~\ref{tab:descriptionmethods_ontology} detail the configuration of the string-based measures and ontology-based measures that are evaluated herein, respectively. Both WBSM and UBSM methods are evaluated in combination with the following word and concept similarity measures: Rada et al. \cite{Rada1989-cv}, Jiang\&Conrath \cite{Jiang1997-zz}, and three state-of-the-art unexplored measures, called cosJ\&C \cite{Lastra-Diaz2015-ct,lastra-diaz2022}, coswJ\&C \cite{Lastra-Diaz2015-ct,lastra-diaz2022}, and Cai et al. \cite{Cai2017-di,lastra-diaz2022}. The word similarity measure which reports the best results is used to evaluate the COM method \cite{Sogancioglu2017-rc,Rada1989-cv}. Table \ref{tab:descriptionmethods_sentenceEmbeddings} details the sentence similarity methods based on the evaluation of pre-trained character, word, and Sentence Embedding (SE) models that are evaluated in this work. Finally, table \ref{tab:descriptionmethods_languageModels} details the pre-trained language models that are evaluated in our experiments.

\begin{table*}[!h]
\caption{Detailed setup for the string-based sentence similarity measures which are evaluated in this work. All the string-based measures follow the implementation of Sogancioglu et al. \cite{Sogancioglu2017-rc}, who use the Simmetrics library \cite{Chapman2005-wx}. LiBlock method proposed herein is an adaptation from Li et al. \cite{Li2006-au} combined with a string-based measure, as detailed in the previous section.}
\captionsetup{width=\textwidth}
{\begin{tabular}{cll}
\hline
ID & Method & Detailed setup of each method  \\
\hline
\\
M1 & Qgram \cite{Ukkonen1992-uf} & \makecell[l]{ $ sim(a,b) = \frac{2 \times \mid q-grams(a)  \cup  q-grams(b) \mid}{\mid q-grams(a) \mid + \mid q-grams(b) \mid} $, being $a$ \\ and $b$ sets of q words, and with q = 3. }\\
\\
M2 & Jaccard \cite{Jaccard1908-hk, Manning1999-ja} & \makecell[l]{ $sim(a,b) = \frac{\mid a \cup b \mid}{\mid a \cap b \mid}$, being $a$ and $b$ sets of words \\ of the first and second sentence respectively. } \\
\\
M3 & \makecell[l]{Block distance \\ \cite{Krause1986-wb}} &  \makecell[l]{ $sim(s_1,s_2) =  1 - \frac{\sum\limits_{i=1}^{|D|} |fr(w_i, s_1) - fr(w_i, s_2)|}{\sum\limits_{i=1}^{|D|}fr(w_i, s_1 + s_2)}$, \\ as detailed in equation \ref{eq:BlockDist}.}\\
\\
M4 & \makecell[l]{LiBlock \\ (this work)} & \makecell[l]{LiBlock method (see eq. \ref{eq:LiBlock}) annotated with \\ CUI concepts and using cTAKES combined \\ with the Block Distance \cite{Krause1986-wb} method using \\ its best pre-processing configuration.} \\
\\
M5 & \makecell[l]{Levenshtein \\ distance \cite{Levenshtein1966-by}} &  \makecell[l]{Measures the minimal cost number of insertions, \\ deletions and replacements needed for  \\ transforming the first into the second sentence. \\ Insert, delete and substitution cost set to 1.}  \\
\\
M6  & \makecell[l]{Overlap \\ coefficient \cite{Lawlor1980-pm}} &  \makecell[l]{$sim(a,b) = \frac{\mid a \cap b \mid}{\mid Min(\mid a \mid, \mid b \mid) \mid}$, being $a$ and $b$ sets of \\ words of the first and second sentence respectively.} \\
\hline
\end{tabular}}{}
\label{tab:descriptionmethods_string}
\end{table*}

\begin{table*}[!h]
\caption{Detailed setup for the ontology-based sentence similarity measures evaluated in this work. The evaluation of the methods using Rada \cite{Rada1989-cv}, coswJ\&C \cite{Lastra-Diaz2015-ct}, and Cai \cite{Cai2017-di} word similarity measures use a reformulation of the original path-based measures based on the new Ancestors-based Shortest-Path Length (AncSPL) algorithm \cite{lastra-diaz2022}.}
\captionsetup{width=\textwidth}
{\begin{tabular}{cll}
\hline
ID & Sentence similarity method & Detailed setup of each method \\
\hline
\\
M7 & WBSM-Rada \cite{Sogancioglu2017-rc, Rada1989-cv,lastra-diaz2022} & \makecell[l]{WBSM \cite{Sogancioglu2017-rc} combined with Rada \cite{Rada1989-cv} \\ measure  using the AncSPL algorithm \cite{lastra-diaz2022} } \\
\\
M8 & WBSM-J\&C \cite{Sogancioglu2017-rc, Jiang1997-zz,Sanchez2011-cf} & \makecell[l]{WBSM \cite{Sogancioglu2017-rc} combined with J\&C \cite{Jiang1997-zz} \\ measure and Sanchez et al. \cite{Sanchez2011-cf} IC model} \\
 \\
M9 & \makecell[l]{WBSM-cosJ\&C \\ \cite{Sogancioglu2017-rc, Lastra-Diaz2015-ct,lastra-diaz2022} (this work) } & \makecell[l]{ WBSM \cite{Sogancioglu2017-rc} with cosJ\&C \cite{Lastra-Diaz2015-ct} \\ measure and Sanchez et al. \cite{Sanchez2011-cf} IC model \\ using the AncSPL algorithm \cite{lastra-diaz2022}} \\
 \\
M10 & \makecell[l]{WBSM-coswJ\&C \\ \cite{Sogancioglu2017-rc, Lastra-Diaz2015-ct,Sanchez2011-cf,lastra-diaz2022}  (this work) } & \makecell[l]{ WBSM \cite{Sogancioglu2017-rc} with coswJ\&C \cite{Lastra-Diaz2015-ct} measure \\ and  Sanchez et al. \cite{Sanchez2011-cf} IC model  \\ using the AncSPL algorithm \cite{lastra-diaz2022}} \\
 \\
M11 & \makecell[l]{WBSM-Cai \cite{Sogancioglu2017-rc, Cai2017-di,lastra-diaz2022}} & \makecell[l]{WBSM \cite{Sogancioglu2017-rc} combined with Cai et al. \cite{Cai2017-di} \\ measure and Cai et al. \cite{Cai2017-di} IC model \\ using the AncSPL algorithm \cite{lastra-diaz2022}} \\
\\
M12 & UBSM-Rada \cite{Sogancioglu2017-rc, Rada1989-cv,lastra-diaz2022} & \makecell[l]{UBSM \cite{Sogancioglu2017-rc} with Rada et al. \cite{Rada1989-cv} \\ measure using the AncSPL algorithm \cite{lastra-diaz2022} } \\
\\
M13 & UBSM-J\&C \cite{Sogancioglu2017-rc, Jiang1997-zz, Sanchez2011-cf} & \makecell[l]{UBSM \cite{Sogancioglu2017-rc} combined with J\&C \cite{Jiang1997-zz} \\ measure and Sanchez et al. \cite{Sanchez2011-cf} IC model} \\
 \\
M14 & \makecell[l]{UBSM-cosJ\&C \\ \cite{Sogancioglu2017-rc, Lastra-Diaz2015-ct,Sanchez2011-cf} (this work) } & \makecell[l]{ UBSM \cite{Sogancioglu2017-rc} with cosJ\&C \cite{Lastra-Diaz2015-ct} measure \\  and Sanchez et al. \cite{Sanchez2011-cf} IC model \\ using the AncSPL algorithm \cite{lastra-diaz2022}}\\
 \\
M15 & \makecell[l]{UBSM-coswJ\&C \\ \cite{Sogancioglu2017-rc, Lastra-Diaz2015-ct,Sanchez2011-cf,lastra-diaz2022} (this work)} & \makecell[l]{ UBSM \cite{Sogancioglu2017-rc} with coswJ\&C \cite{Lastra-Diaz2015-ct} measure \\  and Sanchez et al. \cite{Sanchez2011-cf} IC model \\ using the AncSPL algorithm \cite{lastra-diaz2022}}\\
 \\
M16 & \makecell[l]{UBSM-Cai \cite{Sogancioglu2017-rc, Cai2017-di,lastra-diaz2022} } & \makecell[l]{UBSM \cite{Sogancioglu2017-rc} combined with Cai et al. \cite{Cai2017-di} \\ measure and Cai et al. \cite{Cai2017-di} IC model \\ using the AncSPL algorithm \cite{lastra-diaz2022}} \\
\\
M17 & \makecell[l]{COM \cite{Sogancioglu2017-rc,Rada1989-cv}} & \makecell[l]{$\lambda \cdot$WBSM-Rada + $(1 - \lambda) \cdot$UBSM-Rada \\ with $\lambda=0.5$} \\
\hline
\end{tabular}}{}
\label{tab:descriptionmethods_ontology}
\end{table*}

\begin{table*}[!h]
\caption{Detailed setup for the sentence similarity methods based on pre-trained character, word (WE) and sentence (SE) embedding models evaluated herein.}
\captionsetup{width=\textwidth}
{\begin{tabular}{cll}
\hline
ID & Sentence similarity method & Detailed setup of each method \\
\hline
\\
M18 & \makecell[l]{ Flair \cite{Akbik2018-fh} } & \makecell[l]{ Contextual string embeddings \\ trained on PubMed} \\
M19 & \makecell[l]{ Pyysalo et al. \cite{Pyysalo2013-jy} } & \makecell[l]{ Skip-gram trained on PubMed + PMC} \\
M20 & \makecell[l]{ BioConceptVec \cite{Chen2019-jo} } & \makecell[l]{ Skip-gram WE model trained on PubMed \\ using word2vec program} \\
M21 & \makecell[l]{ BioConceptVec \cite{Chen2019-jo} } & \makecell[l]{ CBOW WE model trained on PubMed \\ using word2vec program} \\
M22 & \makecell[l]{ Newman-Griffis et al.\cite{Newman-Griffis2017-mz}  } & \makecell[l]{ Skip-gram WE model trained on PubMed \\ using word2vec program} \\
M23 & \makecell[l]{ Newman-Griffis et al.\cite{Newman-Griffis2017-mz} } & \makecell[l]{ CBOW WE model trained on PubMed \\ using word2vec program} \\
M24 & \makecell[l]{ Newman-Griffis et al.\cite{Newman-Griffis2017-mz} } & \makecell[l]{ GloVe WE model trained on PubMed} \\
M25 & \makecell[l]{ BioConceptVec$_{GloVe}$ \cite{Chen2019-jo} } & \makecell[l]{GloVe We model trained on PubMed} \\
\\
M26 & BioWordVec$_{int}$ \cite{Zhang2019-qq} & \makecell[l]{FastText \cite{Bojanowski2017-pb} WE model trained on \\ PubMed + MeSH} \\
M27 & BioWordVec$_{ext}$ \cite{Zhang2019-qq} & \makecell[l]{FastText \cite{Bojanowski2017-pb} trained on PubMed + MeSH} \\
\\
M28 & BioNLP2016$_{win2}$ \cite{Chiu2016-bs} & \makecell[l]{FastText \cite{Bojanowski2017-pb} WE model based on skip-gram \\ and trained on PubMed with training setup \\ detailed in \cite[table 18]{Chiu2016-bs}} \\
\\
M29 & BioNLP2016$_{win30}$ \cite{Chiu2016-bs} & \makecell[l]{FastText \cite{Bojanowski2017-pb} WE model  based on skip-gram \\ and trained on PubMed with training setup \\ detailed in \cite[table 18]{Chiu2016-bs}} \\
\\
M30 & \makecell[l]{ BioConceptVec$_{fastText}$ \cite{Chen2019-jo} } & \makecell[l]{FastText \cite{Bojanowski2017-pb} WE model trained on PubMed} \\
\\
M31 & \makecell[l]{ Universal Sentence \\ Encoder (USE) \cite{Cer2018-cr} } & USE SE pre-trained model of Cer et al. \cite{Cer2018-cr} \\
M32 & BioSentVec \cite{Chen2018-uh} & \makecell[l]{sent2vec \cite{Pagliardini2017-og} SE model trained on PubMed \\ + MIMIC-III} \\
\\
M33 & \makecell[l]{FastText-Skipgram-BioC \\ (this work) } & \makecell[l]{FastText \cite{Bojanowski2017-pb} WE model based on Skip-gram \\and trained on PMC-BioC corpus (05,09,2019) \\ with the following setup: vector dim. = 200, \\ learning rate = 0.05, sampling thres. = 1e-4, \\ and negative examples = 10} \\
\hline
\end{tabular}}{}
\label{tab:descriptionmethods_sentenceEmbeddings}
\end{table*}

\begin{table*}[!h]
\caption{Detailed setup for the sentence similarity methods based on pre-trained language models evaluated in this work.}
\captionsetup{width=\textwidth}
{\begin{tabular}{cll}
\hline
ID & Sentence similarity method & Detailed setup of each method  \\
\hline
\\
M34 & \makecell[l]{BioBERT Base 1.0 \cite{Lee2019-pb} \\ (+ PubMed) } & \makecell[l]{ BERT \cite{Devlin2018-eq} trained on English Wikipedia \\ + BooksCorpus + PubMed abstracts}\\
\\
M35 & \makecell[l]{BioBERT Base 1.0 \cite{Lee2019-pb} \\ (+ PMC) } & \makecell[l]{ BERT \cite{Devlin2018-eq} trained on English \\ Wikipedia + \\ BooksCorpus + PMC full-text articles}\\
\\
M36 & \makecell[l]{BioBERT Base 1.0 \cite{Lee2019-pb} \\ (+ PubMed + PMC) } & \makecell[l]{ BERT \cite{Devlin2018-eq} trained on English Wikipedia \\ + BooksCorpus + PubMed \\ abstracts + PMC full-text articles }\\ 
\\
M37 & \makecell[l]{BioBERT Base 1.1 \cite{Lee2019-pb} \\ (+ PubMed) } & \makecell[l]{ BERT \cite{Devlin2018-eq} trained on English Wikipedia \\ + BooksCorpus + PubMed abstracts }\\
\\
M38 & \makecell[l]{BioBERT Large 1.1 \cite{Lee2019-pb}\\ (+ PubMed) } & \makecell[l]{ BERT \cite{Devlin2018-eq} trained on English Wikipedia \\ + BooksCorpus + PubMed abstracts }\\
\\
M39 & \makecell[l]{NCBI-BlueBERT \\ Base \cite{Peng2019-cc} PubMed } & BERT \cite{Devlin2018-eq} trained on PubMed abstracts \\
\\
M40 & \makecell[l]{NCBI-BlueBERT \\ Large \cite{Peng2019-cc} PubMed } & BERT \cite{Devlin2018-eq} trained on PubMed abstracts \\
\\
M41 & \makecell[l]{ NCBI-BlueBERT \\ Base \cite{Peng2019-cc} \\ PubMed + MIMIC-III } & \makecell[l]{ BERT \cite{Devlin2018-eq} trained on PubMed abstracts \\ + MIMIC-III } \\
\\
M42 & \makecell[l]{ NCBI-BlueBERT \\ Large \cite{Peng2019-cc} \\ PubMed + MIMIC-III } & \makecell[l]{ BERT \cite{Devlin2018-eq} trained on PubMed abstracts \\ + MIMIC-III } \\
\\
M43 & SciBERT \cite{Beltagy2019-pq} & BERT \cite{Devlin2018-eq} trained on PubMed abstracts \\
M44 & ClinicalBERT \cite{Huang2019-vq} & BERT \cite{Devlin2018-eq} trained on PubMed abstracts \\
\\
M45 & \makecell[l]{ PubMedBERT \cite{Gu2020-vm} \\ (abstracts)} & BERT \cite{Devlin2018-eq} trained on PubMed abstracts \\
\\
M46 & \makecell[l]{ PubMedBERT \cite{Gu2020-vm} \\ (abstracts + full text)} & \makecell[l]{BERT \cite{Devlin2018-eq} trained on PubMed abstracts \\ + full text} \\
\\
M47 & \makecell[l]{ ouBioBERT-Base \cite{Wada2020-nw} \\ (Uncased)} & BERT \cite{Devlin2018-eq} trained on PubMed abstracts \\
M48 & \makecell[l]{ BioClinicalBERT \cite{Alsentzer2019-hj}} & BERT \cite{Devlin2018-eq} trained on MIMIC-III \\
M49 & \makecell[l]{ BioDischargesummaryBERT \\ \cite{Alsentzer2019-hj} } & BERT \cite{Devlin2018-eq} trained on MIMIC-III summaries \\
M50 & \makecell[l]{ DischargesummaryBERT \cite{Alsentzer2019-hj} } & BERT \cite{Devlin2018-eq} trained on MIMIC-III summaries  \\
\hline
\end{tabular}}{}
\label{tab:descriptionmethods_languageModels}
\end{table*}

\subsubsection*{Pre-processing methods evaluated in this work}

\begin{figure}[h!]
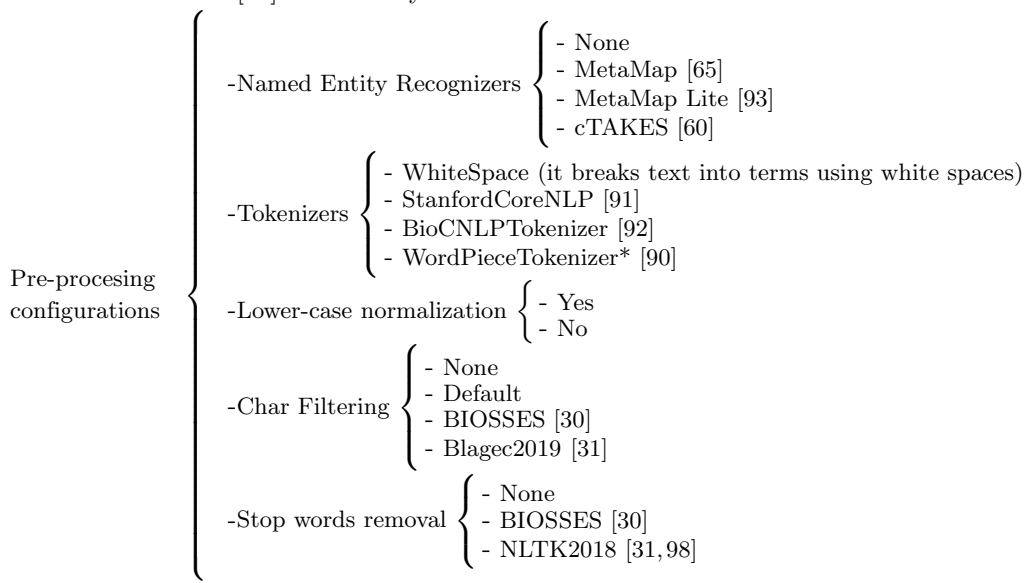

\caption{Detail of the pre-processing configurations that are evaluated in this work. (*) WordPieceTokenizer \cite{Wu2016-en} is used only for BERT-based methods.}
\captionsetup{width=\textwidth}
\AB{\begin{tabular}{l}   \small{Pre-procesing} \\ \small{configurations} \\  \end{tabular}}
{
    \begin{tabular}{l}
        -\AB{\small{Named Entity Recognizers}}
        {
            - \small{None} \\
            - \small{MetaMap \cite{Aronson2010-pb}} \\
            \vspace{0.2em} \\
            - \small{MetaMap Lite \cite{Demner-Fushman2017-zs}} \\
            \vspace{0.2em} \\
            - \small{cTAKES \cite{Savova2010-ed}} \\
        } \\
        -\AB{\small{Tokenizers}}
        {
            - \small{WhiteSpace (it breaks text into terms using white spaces)} \\
            - \small{StanfordCoreNLP \cite{Manning2014-fy}} \\
            \vspace{0.2em} \\
            - \small{BioCNLPTokenizer \cite{Comeau2013-uk}} \\
            \vspace{0.2em} \\
            - \small{WordPieceTokenizer* \cite{Wu2016-en}} \\
        } \\
        -\AB{\small{Lower-case normalization}}
        {
            - \small{Yes} \\
            - \small{No} \\
        } \\
       -\AB{\small{Char Filtering}}
        {
            - \small{None} \\
            - \small{Default} \\
            \vspace{0.3em} \\
            - \small{BIOSSES \cite{Sogancioglu2017-rc}} \\
            \vspace{0.2em} \\
            - \small{Blagec2019 \cite{Blagec2019-nl}} \\
        } \\
        -\AB{\small{Stop words removal}}
        {
            - \small{None} \\
            - \small{BIOSSES \cite{Sogancioglu2017-rc}} \\
            \vspace{0.2em} \\
            - \small{NLTK2018 \cite{Bird2009-ei,Blagec2019-nl}} \\
        } \\
    \end{tabular}
}
\label{fig:pre-processing_combinations}
\end{figure}

The pre-processing stage aims to ensure a fair comparison of the methods that are evaluated in a single end-to-end pipeline. To achieve this later goal, the pre-processing stage normalizes and decomposes the sentences into a series of components that evaluate the same sequence of words applied to all the methods simultaneously. The selection criteria of the pre-processing components have been conditioned by the following constraints: (a) the pre-processing methods and tools used by state-of-the-art methods; and (b) the availability of resources and software tools. Figure \ref{fig:pre-processing_combinations} details all the possible combinations of pre-processing configurations that are evaluated in this work. String, word and sentence embedding, and ontology-based methods, are evaluated using all the available configurations except the WordPieceTokenizer \cite{Wu2016-en}, which is specific to BERT-based methods. Thus, BERT-based methods are evaluated using different char filtering, lower casing normalization, and stop words removal configurations. We use the Pearson and Spearman correlation metrics together with their harmonic score values to determine the impact of the different pre-processing configurations on the performance of the methods evaluated herein. However, we set the best overall performing pre-processing configuration using the harmonic average scores, as well as answering the remaining research questions. 

Most methods receive as input the sequences of words making up the sentences to be compared. The process of splitting sentences into words can be carried out by tokenizers, such as the well-known general domain Stanford CoreNLP tokenizer \cite{Manning2014-fy}, which is used by Blagec et al. \cite{Blagec2019-nl}, or the biomedical domain BioCNLPTokenizer \cite{Comeau2013-uk}. On the other hand, the use of lexicons instead of tokenizers for sentence splitting would be inefficient because of the vast general and biomedical vocabulary. Besides, there would not be possible to provide a fair comparison of the methods because the pre-trained language models have no identical vocabularies.

The tokenized words that conform the sentence, named tokens, are usually pre-processed by removing special characters and lower-casing, and removing the stop words. To analyze all the possible combinations of token pre-processing configurations from the literature, we replicate for each method those pre-processing configurations used by other authors, such as Blagec et al. \cite{Blagec2019-nl} and Sogancioglu et al. \cite{Sogancioglu2017-rc}, and we also evaluate all the pre-processing configurations that have not been evaluated yet. We also study the impact of the pre-processing configurations by not removing special characters and stop words from the tokens, nor normalizing them using lower-casing.

Ontology-based sentence similarity methods estimate the similarity of a sentence by exploiting the 'is-a' relationships between the concepts in an ontology. Therefore, the evaluation of any ontology-based method receives a set of concept-annotated pairs of sentences. The aim of the biomedical NER tools is to recognize automatically biomedical entities in pieces of raw text, such as diseases or drugs. We evaluate the impact of the three more broadly-used biomedical NER tools on the performance of the sentence similarity methods, as follows: (a) MetaMap \cite{Aronson2010-pb}, (b) cTAKES \cite{Savova2010-ed}, and (c) MetaMap Lite \cite{Demner-Fushman2017-zs}. MetaMap tool \cite{Aronson2010-pb} is used by UBSM and COM methods \cite{Sogancioglu2017-rc} for recognizing Unified Medical Language System (UMLS) \cite{Bodenreider2004-ec} concepts in the sentences, which is the standard compendium of biomedical vocabularies. Likewise, we use the default configuration of MetaMap restricted to the UMLS sources of SNOMED-CT and MeSH implemented by HESML V1R5 \cite{lastra-diaz2022, 1RRAWJ_2020}, which is defined by the following features: (i) the use of all available semantic types; (ii) the MedPost Part-of-speech tagger \cite{Smith2004-vi}; and (iii) the MetaMap Word-Sense Disambiguation (WSD) module. We also evaluate cTAKES \cite{Savova2010-ed} because it has shown to be a robust and reliable tool to recognize biomedical entities \cite{Reategui2018-kf}. Encouraged by the high computational cost of MetaMap in evaluating large text corpus, Demner-Fushman et al. \cite{Demner-Fushman2017-zs} introduce a lighter MetaMap version, called Metamap Lite, which provides a real-time implementation of the basic MetaMap annotation capabilities without a large degradation of its performance.

Due to the large number of possible combinations of each pre-processing dimension, such as Named Entity Recognizers, tokenizers or char filtering methods, we have evaluated the pre-processing combinations of each dimension by defining a fixed pre-processing configuration for the rest of dimensions, except for the string-based methods, whose performance is high enough to not cause a significant variation in the running time of the experiments.

\subsubsection*{Detailed workflow of our experiments}

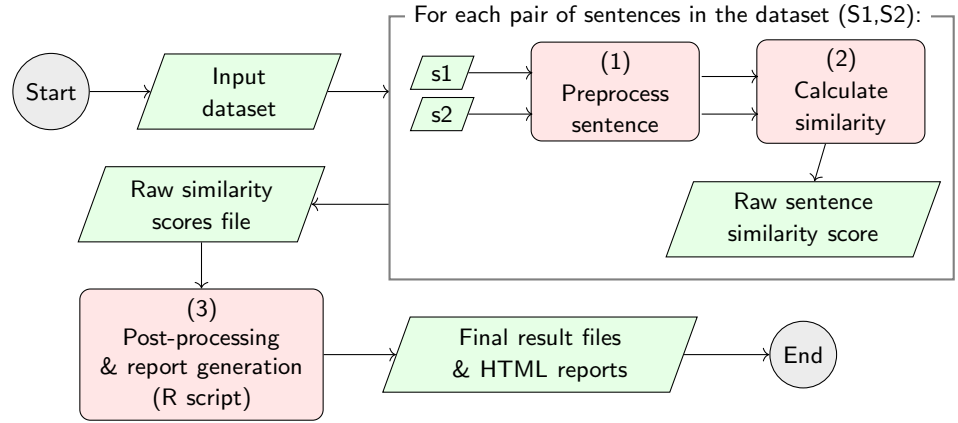
\begin{figure}[h!]
\caption{Detailed workflow implemented by our experiments for pre-processing the input sentences, calculating the raw similarity scores, and post-processing the results obtained in the evaluation of the biomedical datasets. This workflow generates a collection of raw and processed data files.}
\centering
\tikzstyle {process} = [rectangle, rounded corners, draw=black, fill=red!10,
                        minimum width=1cm, minimum height=1cm,text centered, font=\sffamily]
\tikzstyle {outputdata} = [fill=green!10, trapezium, trapezium left angle=70,
                        trapezium right angle=-70, draw=black,
                        minimum width=0.1cm, minimum height=1cm,
                        text centered, font=\sffamily]
\tikzstyle {init} = [circle, draw=black, text centered, font=\sffamily, fill=gray!15]
\tikzstyle {decision} = [diamond, draw=black, text centered, font=\sffamily, fill=blue!10]
\tikzstyle {initSents} = [circle, draw=black, fill=green!10, text centered, font=\sffamily, trapezium, trapezium left angle=70, trapezium right angle=-70]

\begin{tikzpicture}[every node/.style={fill=white, font=\sffamily}, align=center]

\node (start) [init] {\small{Start}};
\node (inputDataset) [outputdata, node distance=2.5cm,text width=1.5cm, right of=start] {\small{Input dataset}};
\node (RS1) at (5.2,0.25) [initSents] {\small{s1}};
\node (RS2) at (5.2,-0.3) [initSents] {\small{s2}};
\node (preprocess) at (7.5,0) [process,text width=2cm] {\small{(1)} \\ \small{Preprocess sentence}};
\node (calculateSim) at (10.5,0) [process,text width=2cm] {\small{(2)} \\ \small{Calculate similarity}};
\node (similarityScore) at (10,-1.7) [outputdata, text width=2.5cm] {\small{Raw sentence} \\ \small{similarity score}};
\node (rawSimilarityScores) at (2,-1.5) [outputdata, text width=2cm] {\small{Raw similarity} \\ \small{scores file}};
\node (postProcessing) [process, node distance=2cm,text width=3cm, below of=rawSimilarityScores] {\small{(3)} \\ \small{Post-processing} \\ \small{\& report generation} \\ \small{(R script)}};
\node (processedSimilarityScores) [outputdata, node distance=4.5cm,text width=3cm,right of=postProcessing] {\small{Final result files} \\ \small{\& HTML reports}};
\node (finish) [init, node distance=3.5cm, right of=processedSimilarityScores] {\small{End}};

\draw[->] (start) -- (inputDataset);
\draw[->] (inputDataset) -- (4.5,0);
\draw[->] (RS1) -- (6.4,0.25);
\draw[->] (RS2) -- (6.4,-0.3);
\draw[->] (8.65,0.2) -- (9.4,0.2);
\draw[->] (8.65,-0.3) -- (9.4,-0.3);
\draw[->] (calculateSim) -- (similarityScore);
\draw[->] (4.5,-1.5) -- (rawSimilarityScores);
\draw[->] (rawSimilarityScores) -- (postProcessing);  
\draw[->] (postProcessing) -- (processedSimilarityScores);
\draw[->] (processedSimilarityScores) -- (finish);    

\node at (9,0.37) [fill=white,text opacity=1,fill opacity=0] {\scriptsize{S1}};
\node at (9,-0.15) [fill=white,text opacity=1,fill opacity=0] {\scriptsize{S2}};

\draw [color=gray,thick](4.5,1) rectangle (12,-2.5);
\node at (4.7,1) [above=5mm, right=0mm] {\small{For each pair of sentences in the dataset (S1,S2):}};

\end{tikzpicture}
\label{fig:reproducibility_workflow}
\end{figure}

\begin{figure}[h!]
\caption{Detailed sentence pre-processing workflow that are implemented in our experiments. The pre-processing stage takes an input sentence and produces a pre-processed sentence as output. (*) The named entity recognizer are only evaluated in ontology-based methods.}
\centering
\tikzstyle {process} = [rectangle, rounded corners, draw=black, fill=red!10,
                        minimum width=1cm, minimum height=1cm,text centered, font=\sffamily]
\tikzstyle {outputdata} = [fill=green!10, trapezium, trapezium left angle=70,
                        trapezium right angle=-70, draw=black,
                        minimum width=1cm, minimum height=1cm,
                        text centered, font=\sffamily]
\tikzstyle {init} = [circle, draw=black, text centered, font=\sffamily, fill=gray!15]
\tikzstyle {decision} = [diamond, draw=black, text centered, font=\sffamily, fill=blue!10]

\begin{tikzpicture}[every node/.style={fill=white, font=\sffamily}, align=center]

\node (start) [init] {\small{Start}};
\node (RS) [outputdata, node distance=2.5cm,text width=1.5cm, right of=start] {\small{Raw sentence}};
\node (ner) [process, node distance=3.5cm,text width=2cm, right of=RS] {\small{(1.a)} \\ \small{NER*}};
\node (tokenizer) [process, node distance=3.5cm,text width=2cm, right of=ner] {\small{(1.b) Tokenizer}};
\node (lowercaseNormalization) [process, node distance=2cm,text width=2cm,below of=tokenizer] {\small{(1.c) Lower-case normalization}};
\node (charFiltering) [process, node distance=3cm,text width=2cm,left of=lowercaseNormalization] {\small{(1.d)} \\ \small{Char filtering}};
\node (stopWordsRemoval) [process, node distance=3cm,text width=2cm,left of=charFiltering] {\small{(1.e)} \\ \small{Stop words removal}};
\node (preprocessedSentence) [outputdata, node distance=3cm,text width=2cm,left of=stopWordsRemoval] {\small{Preprocessed sentence}};
\node (finish) [init, node distance=2.5cm, left of=preprocessedSentence] {\small{End}};

\draw[->] (start) -- (RS);
  \draw[->] (RS) -- (ner);
  \draw[->] (ner) -- (tokenizer);
  \draw[->] (tokenizer) -- (lowercaseNormalization);
  \draw[->] (lowercaseNormalization) -- (charFiltering);
  \draw[->] (charFiltering) -- (stopWordsRemoval);
  \draw[->] (stopWordsRemoval) -- (preprocessedSentence);
  \draw[->] (preprocessedSentence) -- (finish);    

\end{tikzpicture}
\label{fig:pre-processing_workflow}
\end{figure}
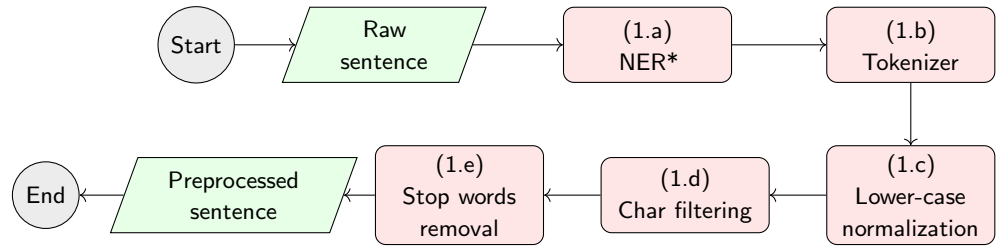

Figure \ref{fig:reproducibility_workflow} shows the workflow for running the experiments implemented in this work. Given an input dataset, such as BIOSSES \cite{Sogancioglu2017-rc}, MedSTS \cite{Wang2018-oj}, or CTR \cite{Lithgow-Serrano2019-si}, the first step is to pre-process all the sentences, as shown in figure \ref{fig:pre-processing_workflow}. For each sentence pair $(s_1,s_2)$ in the dataset, the pre-processing stage is divided into four stages as follows: (1.a) named entity recognition of UMLS \cite{Bodenreider2004-ec} concepts, using different state-of-the-art NER tools, such as MetaMap \cite{Aronson2010-pb} or cTAKES \cite{Savova2010-ed}; (1.b) tokenization of the sentences, using well-known tokenizers, such as the Stanford CoreNLP tokenizer \cite{Manning2014-fy}, BioCNLPTokenizer \cite{Comeau2013-uk}, or WordPieceTokenizer \cite{Wu2016-en} for BERT-based methods; (1.c) lower-case normalization; (1.d) character filtering, which allows the removal of punctuation marks or special characters; and finally, (1.e) the removal of stop-words, following different approximations evaluated by other authors like Blagec et al. \cite{Blagec2019-nl} or Sogancioglu et al. \cite{Sogancioglu2017-rc}. Once each dataset is pre-processed in step 1 detailed in figure \ref{fig:reproducibility_workflow}), the aim of step 2 is to calculate the similarity score between each pair of sentences in the dataset to produce a raw output file containing all raw similarity scores, one score per sentence pair. Finally, a R-language script is used in step 3 to process the raw similarity files and produce the final human-readable tables reporting the Pearson and Spearman correlation values shown in table \ref{tab:table_results}, as well as the statistical significance of the results and any other supplementary data table required by our study on the impact of the pre-processing and NER tools reported in appendices A and B respectively.

Finally, we also evaluate all the pre-processing combinations for each family of methods to study the impact of the pre-processing methods on the performance of the sentence similarity methods, with the only exception of the BERT-based methods. The pre-processing configurations of the BERT-based methods are only evaluated in combination with the WordPiece Tokenizer \cite{Wu2016-en} because it is required by the current BERT implementations.


\subsubsection*{Evaluation metrics}

The evaluation metrics used to compare the performance of the methods analyzed are the following: (1) the Pearson correlation, denoted by $r$ in equation (\ref{eq_Pearson}); (2) the Spearman rank correlation, denoted by $\rho $ in equation (\ref{eq_Spearman}); (3) and the harmonic score, denoted by $h$ in equation (\ref{eq_Harmonic}). The Pearson correlation evaluates the linear correlation between two random samples, whilst the Spearman rank correlation is rank-invariant and evaluates the monotonic relationship between two random samples, and the harmonic score allows comparing sentence similarity methods by using a single weighted score based on their performance in Pearson and Spearman correlation.

\begin{eqnarray}
r &=&\frac{\sum\nolimits_{i=1}^{n}\left( X_{i}-\overline{X}\right) \left(
Y_{i}-\overline{Y}\right) }{\sqrt{\sum\nolimits_{i=1}^{n}\left( X_{i}-
\overline{X}\right) ^{2}}\sqrt{\sum\nolimits_{i=1}^{n}\left( Y_{i}-
\overline{Y}\right) ^{2}}}  \label{eq_Pearson} \\
\rho &=&1-\frac{6\sum\nolimits_{i=1}^{n}d_{i}^{2}}{n\left( n^{2}-1\right) },
\quad \quad \quad d{i}=\left( x_{i}-y_{i}\right)  \label{eq_Spearman} \\
h &=&\frac{2r\rho}{r+\rho} \label{eq_Harmonic}
\end{eqnarray}

Finally, we use the well-known t-Student test to carry-out a statistical significance analysis of the results of the evaluation of the methods in the tree biomedical datasets shown in table \ref{tab:table_Datasets}. In order to compare the overall performance of the semantic measures that is evaluated in our experiments, we use the harmonic score average in all datasets. The statistical significance of the results is evaluated using the p-values resulting from the t-student test for the mean difference between the harmonic score values reported by each pair of semantic measures in all datasets. The p-values are computed using a one-sided t-student distribution on two paired random sample vectors made up by the harmonic ($h$) score values obtained in the evaluation of the three aforementioned datasets. Our null hypothesis, denoted by $H_0$, is that the difference in the average performance between each pair of compared sentence similarity methods is 0, whilst the alternative hypothesis, denoted by $H_1$, is that their average performance is different. For a 5\% level of significance, it means that if the p-value is greater or equal than 0.05, we must accept the null hypothesis. Otherwise, we can reject $H_0$ with an error probability of less than the p-value. In this latter case, we say that a first sentence similarity method obtains a statistically significantly higher value than the second one or that the former one significantly outperforms the second one.

\paragraph{Uniform size datasets for our statistical significance analysis.} The scarcity of the datasets and the notable size difference among datasets varying from 100 to 1,068 sentence pairs prevent both from studying the statistical significance of the results with adequate sample size and carry-out a fair comparison of the results. For this reason, we have divided the MedSTS dataset into 10 parts considered as independent datasets to perform the study of the statistical significance of the results. Thus, we have artificially obtained 12 datasets of 100 to 200 pairs of sentences. This set of datasets allows us to obtain the p-values comparing the statistical significance between the measure, but does not modify the processed results from table \ref{tab:table_results}. All the necessary resources for obtaining both the table \ref{tab:table_results} and the table containing all the p-values reported in Appendix A are publicly available in the reproducibility dataset and the companion Lab Protocol article under preparation, as detailed in table \ref{tab:table_material}.

\subsubsection*{Statistical performance analysis of the best methods}

In order to answer the RQ5 research question, we study how well the sentence similarity methods are estimating the degree of semantic similarity between two sentences by analyzing the deviation of their estimated values regarding the human similarity scores. We want to analyze why the methods are doing well or bad on specific sentence pairs to elucidate some explanation to this behaviour, as well as identifying the main drawbacks and limitations of the current state-of-the-art methods. To carry out this performance analysis, we analyze the statistics of the similarity error function $E_{sim}$ of the methods defined in equation \ref{eq_simerror}. We only use some sentences extracted from the BIOSSES dataset for this analysis because this dataset has no licensing restrictions on its use, which allows us to reproduce their sentences herein, unlike MedSTS. On the other hand, we could have also used CTR because it has no licensing restrictions; however, CTR has not been previously used in this sentence similarity task.

\begin{align}
E_{sim} &: L_\Sigma\times L_\Sigma \rightarrow [0,1] \subset \mathbb{R} \nonumber \\
E_{sim}(s_1,s_2) &= sim(s_1,s_2) - humanSim(s_1,s_2) \label{eq_simerror}
\end{align}

Our methodology to conduct the performance analysis is detailed below:
\begin{itemize}
    \item[1.]  Selection of the best-performing method from each family of methods.
    \item[2.]  Estimation of the Probability Density Function (PDF) of the $E_{sim}$ function for the evaluation of the selected best-performing methods in each dataset by calling the ``$density$" function provided by the R statistical package.
    \item[3.] Selection of the sentences based on their similarity error in the BIOSSES dataset:
    \begin{itemize}
        \item[3.1] the sentences with the lowest and highest absolute similarity error $|E_{sim}|$ for each method are extracted.
        \item[3.2] each sentence selected in the step above is pre-processed using the best pre-processing configuration for each method.
        \item[3.3] the resulting pre-processed sentences and the statistical information of the similarity scores are analyzed in the \emph{Discussion} section.
    \end{itemize}
\end{itemize}

\subsubsection*{Software implementation}

We have developed a new sentence measures library for the biomedical domain called HESML-STS, which is based on HESML V1R5 \cite{lastra-diaz2022,Lastra-Diaz2017-qo}, as detailed in table \ref{tab:table_material}. All our experiments are generated by running the \emph{HESMLSTSclient} and \emph{HESMLSTSImpactpre-processingclient} programs, which generates a raw output file in comma-separated file format (*.csv) for each dataset detailed in table \ref{tab:table_Datasets}. The raw output files contain the raw similarity values returned by each sentence similarity method in the evaluation of the degree of similarity between sentences. The final results for the Pearson and Spearman correlation, and the harmonic values detailed in table \ref{tab:table_results} are automatically generated by running a R-language script file on the collection of raw similarity files, which also generates all the tables reported in appendices A and B provided as supplementary material. All tables are written both in Latex and comma-separated file format (*.csv) formats. For a more detailed description of the protocol for running our experiments, we refer the reader to the protocol \cite{Lara-Clares2022protocolsIO} detailed in appendix C.

We implemented a parser for loading pre-trained embedding models based on FastText \cite{Bojanowski2017-pb} and other word embedding models \cite{Pyysalo2013-jy,Chen2019-jo,Newman-Griffis2017-mz,Zhang2019-qq,Chiu2016-bs}, which are efficiently evaluated as sentence similarity measures in HESML by implementing the averaging Simple Word EMbeddings (SWEM) approach introduced by Shen et al. \cite{Shen2018-ky}. On the other hand, the software replication required to evaluate sentence embeddings and BERT-based language models is extremely complex and out of the scope of this work. For this reason, these models are evaluated using the original software artifacts used to generate the aforementioned pre-trained models. Thus, we implemented a collection of Python wrappers for evaluating the available models by using the provided software artifacts as follows: (1) Sent2vec-based models \cite{Chen2018-uh} are evaluated using the Sent2vec library \cite{Pagliardini2017-og}; (2) Flair models \cite{Akbik2018-fh} are evaluated using the flairNLP framework \cite{Akbik2018-fh}; and USE models \cite{Cer2018-cr} are evaluated using the open source platform TensorFlow \cite{Abadi2016-or}. All BERT-based pre-trained models are evaluated using the open source bert-as-a-service library \cite{xiao2018bertservice}.

\subsubsection*{Reproducing our benchmarks}

For the sake of reproducibility, we introduce a detailed reproducibility protocol at protocols.io \cite{Lara-Clares2022protocolsIO} that is based on a reproducibility dataset \cite{EPNXTR_2021Dataset} containing all the software and data necessary to allow the exact replication of all our experiments and results. Our reproducibility protocol is mainly based on a Docker-based image \footnote{\url{https://hub.docker.com/repository/docker/alicialara/hesml_v2r1}} that include a pre-installation of all the necessary software and the Java source code and binary files of our benchmark program. Our source code files are tagged in Github with a permanent tag named ``SentenceSimilarityBenchmark" \footnote{\url{https://github.com/jjlastra/HESML/releases/tag/Release_HESML_V1R5.0.2}}.

In addition, we plan to submit a Lab Protocol\footnote{\url{https://collections.plos.org/collection/lab-protocols}} article under preparation \cite{Lara-Clares2022LabProtocol}, which will provide a detailed description of the publicly available reproducibility dataset \cite{EPNXTR_2021Dataset} and a very detailed reproduciblility protocol \cite{Lara-Clares2022protocolsIO} to allow the exact replication of all our methods, experiments, and results. We also plan to submit a research article under preparation \cite{HESMLSTSpaper} to introduce the new HESML-STS software library integrated into the latest HESML V2R1 version, together with a set of reproducible benchmarks on semantic measures libraries for biomedical sentence similarity. The new HESML V2R1 release will make publicly available soon, once we have appropriately separated the configurations requiring software restricted by third-party licenses, such as cTAKES and Metamap NER tools, from the rest of the project. However, our reproducibility dataset allows the full and exact replication of all our experiments by completing the licensing requirements of the UMLS databases and the aforementioned NER tools for the National Library of Medicine (NLM) of the United States \footnote{\url{https://www.nlm.nih.gov/databases/umls.html\#license_request}} .

Table \ref{tab:table_material} details all the reproducibility resources provided as supplementary material with this work. Our benchmarks are implemented using Java 8, Python 3 and R programming languages, and thus, they can be reproduced in any Java-complaint or Docker-complaint platforms, such as Windows, MacOS, or any Linux-based system. 

\begin{table}[!ht]
\caption{Supplementary material and reproducibility resources of this work.}
\begin{tabular}{ll}
\small{Material} & \small{Description} \\
\hline
\\
\small{Reproducibility dataset \cite{EPNXTR_2021Dataset}} & \small{\makecell[l]{All raw input and output data files, pre-trained \\ model files, and a long-term reproducibility image \\ based on Docker, which is publicly  available in the \\ Spanish Dataverse Network \footnote{\url{https://doi.org/10.21950/EPNXTR}}}} \\
\\
\small{Reproducibility protocol \cite{Lara-Clares2022protocolsIO}} & \small{\makecell[l]{Raw step-by-step instructions to download the \\ required resources and reproduce the experiments \\  evaluated in this work}} \\
\\
\small{\makecell[l]{Lab Protocol article \cite{Lara-Clares2022LabProtocol} \\ (under preparation)}} & \small{\makecell[l]{Data and methods article introducing a very detailed \\ description of our experiments, datasets, and \\ reproducibility  protocol to allow the independent \\ replication  of our experiments and results}} \\
\\
\small{\makecell[l]{HESML-STS software library \\ (integrated into HESML V2R1)}} & \small{\makecell[l]{Release of the new HESML-STS library. This library \\ is based on the previous HESML V1R5 version \cite{lastra-diaz2022,Lastra-Diaz2017-qo} \\ published in Github \footnote{\url{https://github.com/jjlastra/HESML/tree/HESML-STS\_master\_dev}} and the Spanish Dataverse \\ Network \cite{EPNXTR_2021Dataset} under a CC By-NC-SA-4.0 license.}} \\
\\
\small{\makecell[l]{HESML V2R1 software release \\ (under preparation)}} & \small{\makecell[l]{Release of the new HESML V2R1 version which \\ will be published soon. This new release will be \\ based on the previous HESML V1R5 version, \\ including the new HESML-STS software package \\ that has been developed for this work, after \\ managing all the licensing restrictions of \\ the NER tools.}} \\
\\
\small{\makecell[l]{HESML-STS software paper \cite{HESMLSTSpaper} \\ (under preparation)}} & \small{\makecell[l]{Software article introducing our sentence similarity \\ library, called HESML-STS, together with some \\  benchmarks under preparation.}} \\
\end{tabular}
\label{tab:table_material}
\end{table}

\clearpage

\subsubsection*{Results obtained}
\label{sec:results}

Table \ref{tab:table_pre-processing_methods_selected} shows the selected pre-processing configuration of each method for obtaining their best-performing results, whilst table \ref{tab:table_results} shows the results obtained in the evaluation of all methods in the three biomedical datasets evaluated herein by using their best pre-processing configurations. Table \ref{tab:table_comparison_best_worst_prepro} shows the comparison of results for the highest (best) and lowest (worst) average harmonic score values for the best-performing method of each family shown in blue in table \ref{tab:table_results}, which are defined by the method obtaining the highest average harmonic score. Furthermore, table \ref{tab:ner_comparison} shows the results obtained in our study on the impact of NER tools on the performance of the sentence similarity methods in the evaluation of the MedSTS dataset \cite{Wang2018-oj}. Table \ref{tab:pvalues_ner} shows the harmonic and average harmonic scores obtained in the evaluation of the three biomedical datasets, as well as the resulting p-values comparing the NER tools for each ontology-based method. Table \ref{tab:table_results_liblock} shows the results obtained in the evaluation of the LiBlock method in the three biomedical datasets by using its best pre-processing configuration, and annotating the sentences with all the NER tools combinations. In addition, the aforementioned table details the resulting p-values comparing the best-performing LiBlock-NER combination with the other NER tools. Tables \ref{tab:raw_best_worst_sentences_string}, \ref{tab:raw_best_worst_sentences_ont}, \ref{tab:raw_best_worst_sentences_embedd}, and \ref{tab:raw_best_worst_sentences_bert} show the raw input sentence pairs and their corresponding pre-processed versions in which the best-performing methods obtain the lowest and highest similarity error ($E_{sim}$) in the BIOSSES dataset \cite{Sogancioglu2017-rc}. Table \ref{tab:statsmethods} detail the statistical information for the best-performing methods of each family in the evaluation of the three biomedical datasets evaluated herein. Finally, figure \ref{fig:probabilityerrordistribution} shows the Probability Density Function (PDF) of the similarity error obtained by the best-performing methods of each family in the evaluation of the BIOSSES, MedSTS, and CTR datasets respectively.

On the other hand, appendix A shows the resulting p-values comparing all the methods using their best pre-processing configuration as detailed in \ref{tab:table_results}, which allows us to study the statistical significance of the results, as detailed in the Discussion section. In addition, appendix B shows the experimental results on the impact of pre-processing configurations in all the methods evaluated herein, whose best configuration has been used to determine the final scores for each method. Finally, appendix C detail the protocol for reproducing all the experiments evaluated herein, which is also published in protocols.io \cite{Lara-Clares2022protocolsIO}.

\begin{table*}[!h]
\begin{adjustwidth}{-2.25in}{0in} 
\caption{Best-performing pre-processing configurations used to evaluate the methods compared in this work as reported in table \ref{tab:table_results}, which are derived from our cross-evaluation of each method with the pre-processing configurations shown in figure \ref{fig:pre-processing_combinations} (see Appendix B). (*) COM (M17) uses the best configuration of the WBSM-Rada (M7) and UBSM-Rada (M12) methods for computing the similarity scores.}
{\small \begin{tabular}{@{}llccccc@{}}
\hline
ID & Sentence similarity method & NER & Tokenizer & \makecell[l]{Lower-case} & \makecell[l]{Char \\ filtering} & \makecell[l]{Stop words \\ removal} \\
\hline
M1 & Qgram & None & WhiteSpace & yes & BIOSSES & NLTK2018 \\
M2 & Jaccard & None & WhiteSpace & yes & BIOSSES & NLTK2018 \\
M3 & Block distance & None & WhiteSpace & yes & BIOSSES & NLTK2018 \\
M4 & \makecell[l]{LiBlock (this work)} & cTakes & CoreNLP & yes & Default & NLTK2018 \\
M5 & Levenshtein distance & None & WhiteSpace & no & None & BIOSSES \\
M6 & Overlap coefficient & None & CoreNLP & yes & Default & NLTK2018 \\
\hline
M7 & WBSM-Rada & \makecell[l]{Exact matching} & CoreNLP & yes & BIOSSES & NLTK2018 \\
M8 & WBSM-J\&C & \makecell[l]{Exact matching} & CoreNLP & yes & BIOSSES & None \\
M9 & \makecell[l]{WBSM-cosJ\&C (this work)} & \makecell[l]{Exact matching} & CoreNLP & yes & BIOSSES & None \\
M10 & \makecell[l]{WBSM-coswJ\&C (this work)} & \makecell[l]{Exact matching} & CoreNLP & yes & BIOSSES & NLTK2018 \\
M11 & WBSM-Cai & \makecell[l]{Exact matching} & CoreNLP & yes & BIOSSES & None \\
M12 & UBSM-Rada & cTAKES & CoreNLP & yes & BIOSSES & NLTK2018 \\
M13 & UBSM-J\&C & MetamapLite  & CoreNLP & yes & BIOSSES & NLTK2018 \\
M14 & \makecell[l]{UBSM-cosJ\&C (this work)} & MetamapLite  & CoreNLP & yes & BIOSSES & NLTK2018 \\
M15 & \makecell[l]{UBSM-coswJ\&C (this work)} & cTAKES  & CoreNLP & yes & BIOSSES & NLTK2018 \\
M16 & UBSM-Cai & MetamapLite  & CoreNLP & yes & BIOSSES & NLTK2018 \\
M17 & \makecell[l]{COM (*)}  & - & - & - & - &  \\
\hline
M18 & Flair & None & WhiteSpace & no & BIOSSES & None \\
M19 & Pyysalo et al. & None & CoreNLP & yes & Default & BIOSSES \\
M20 & BioConceptVec$_{word2vec\_sg}$ & None & CoreNLP & yes & Default & BIOSSES \\
M21 & BioConceptVec$_{word2vec\_cbow}$ & None & CoreNLP & yes & Default & BIOSSES \\
M22 & Newman-Griffis$_{word2vec\_sgns}$ & None & CoreNLP & yes & Default & NLTK2018 \\
M23 & Newman-Griffis$_{word2vec\_cbow}$ & None & CoreNLP & yes & Default & NLTK2018 \\
M24 & Newman-Griffis$_{glove}$ & None & CoreNLP & yes & Default & NLTK2018 \\
M25 & BioConceptVec$_{glove}$ & None & CoreNLP & yes & Default & BIOSSES \\
M26 & BioWordVec$_{int}$ & None & CoreNLP & yes & BIOSSES & None \\
M27 & BioWordVec$_{ext}$ & None & CoreNLP & yes & BIOSSES & None \\
M28 & BioNLP2016$_{win2}$ & None & CoreNLP & no & Default & NLTK2018 \\
M29 & BioNLP2016$_{win30}$ & None & CoreNLP & no & Default & NLTK2018 \\
M30 & BioConceptVec$_{fastText}$ & None & CoreNLP & yes & Default & BIOSSES \\
M31 & USE & None & CoreNLP & no & Default & None \\
M32 & \makecell[l]{BioSentVec \\ (PubMed+MIMIC-III) } & None & CoreNLP & yes & BIOSSES & BIOSSES \\
M33 & FastText-SkGr-BioC (this work) & None & CoreNLP & yes & Default & None \\
\hline
M34 & BioBERT Base 1.0 (+ PubMed) & None & WordPiece & yes & BIOSSES & None \\
M35 & BioBERT Base 1.0 (+ PMC) & None & WordPiece & yes & BIOSSES & None \\
M36 & \makecell[l]{BioBERT Base 1.0 (PubMed+PMC)} & None & WordPiece & yes & BIOSSES & None \\
M37 & BioBERT Base 1.1 (+ PubMed) & None & WordPiece& no & Blagec2019 & NLTK2018 \\
M38 & BioBERT Large 1.1 (+ PubMed) & None & WordPiece & no & Blagec2019 & NLTK2018 \\
M39 & NCBI-BlueBERT Base PubMed & None & WordPiece & yes & Blagec2019 & None \\
M40 & NCBI-BlueBERT Large PubMed & None & WordPiece & yes & BIOSSES & None \\
M41 & \makecell[l]{NCBI-BlueBERT \\ Base PubMed + MIMIC-III} & None & WordPiece & yes & BIOSSES & BIOSSES \\
M42 & \makecell[l]{NCBI-BlueBERT \\ Large PubMed + MIMIC-III} & None & WordPiece & yes & BIOSSES & None \\
M43 & SciBERT & None & WordPiece & yes & BIOSSES & NLTK2018 \\
M44 & ClinicalBERT & None & WordPiece & no & Blagec2019 & BIOSSES \\
M45 & PubMedBERT (abstracts) & None & WordPiece  & yes & Default & NLTK2018 \\
M46 & \makecell[l]{PubMedBERT (abstracts+full text)} & None & WordPiece & yes & Default & NLTK2018 \\
M47 & ouBioBERT-Base, Uncased & None & WordPiece & yes & Default & None \\ 
M48 & BioClinicalBERT & None & WordPiece & yes & Blagec2019 & BIOSSES \\
M49 & BioDischargesummaryBERT & None & WordPiece & no & Blagec2019 & NLTK2018 \\ 
M50 & DischargesummaryBERT & None & WordPiece & no & Blagec2019 & NLTK2018 \\
\hline
\end{tabular}}{}
\label{tab:table_pre-processing_methods_selected}
\end{adjustwidth}
\end{table*}

\begin{table*}[!h]
\begin{adjustwidth}{-2.25in}{0in} 
\caption{Pearson (r), Spearman ($\rho$), harmonic ($h$), and harmonic average (AVG) scores obtained by each sentence similarity method evaluated herein in the three biomedical sentence similarity benchmarks arranged by families. All reported values were obtained using the best pre-processing configurations detailed in table \ref{tab:table_pre-processing_methods_selected}. The results in bold show the best scores whilst results in \textcolor{blue}{blue} color show the best average harmonic score for each family.}
{\small \begin{tabular}{@{}llccccccccccccc@{}}
\hline
 & & \multicolumn{3}{c}{\small{BIOSSES} \cite{Sogancioglu2017-rc}} & \multicolumn{3}{c}{\small{MedSTS$_{full}$} \cite{Wang2018-oj}} & \multicolumn{3}{c}{\small{CTR} \cite{Lithgow-Serrano2019-si}} & AVG \\
ID & Sentence similarity methods & r & $\rho$ & h & r & $\rho$ & h & r & $\rho$ & h & h\\
\hline
M1 & Qgram & 0.752 & 0.773 & 0.763 & 0.701 & 0.674 & 0.687 & 0.763 & 0.766 & 0.764 & 0.738 \\
M2 & Jaccard  & 0.782 & 0.815 & 0.798 & 0.706 & 0.680 & 0.693 & 0.759 & 0.797 & 0.777 & 0.756 \\
M3 & Block distance & 0.798 & 0.818 & 0.808 & 0.731 & 0.683 & 0.706 & 0.797 & 0.801 & 0.799 & 0.771 \\
M4 & \makecell[l]{LiBlock (this work)} & 0.820 & \textbf{0.828} & \textbf{0.824} & 0.769 & \textbf{0.710} & 0.739 & 0.793 & 0.808 & 0.800 & \textcolor{blue}{\textbf{0.788}} \\ 
M5 & Levenshtein distance  & 0.529 & 0.536 & 0.533 & 0.610 & 0.634 & 0.622 & 0.498 & 0.536 & 0.516 & 0.557 \\
M6 & Overlap coefficient  & 0.782 & 0.795 & 0.788 & 0.696 & 0.564 & 0.623 & 0.781 & 0.793 & 0.787 & 0.733 \\
\hline
M7 & WBSM-Rada & 0.772 & 0.791 & 0.782 & \textbf{0.774} & 0.709 & \textbf{0.740} & 0.785 & 0.765 & 0.775 & 0.766 \\ 
M8 & WBSM-J\&C & 0.483 & 0.549 & 0.514 & 0.647 & 0.614 & 0.630 & 0.536 & 0.516 & 0.526 & 0.557 \\ 
M9 & WBSM-cosJ\&C (this work) & 0.483 & 0.549 & 0.514 & 0.647 & 0.614 & 0.630 & 0.536 & 0.516 & 0.526 & 0.557 \\
M10 & WBSM-coswJ\&C (this work) & 0.571 & 0.566 & 0.568 & 0.705 & 0.651 & 0.677 & 0.637 & 0.590 & 0.613 & 0.619 \\ 
M11 & WBSM-Cai & 0.458 & 0.542 & 0.497 & 0.629 & 0.601 & 0.615 & 0.492 & 0.459 & 0.475 & 0.529 \\ 
M12 & UBSM-Rada & 0.792 & 0.809 & 0.800 & 0.763 & 0.700 & 0.730 & 0.776 & 0.794 & 0.785 & 0.772 \\  
M13 & UBSM-J\&C & 0.529 & 0.573 & 0.550 & 0.683 & 0.621 & 0.650 & 0.620 & 0.585 & 0.602 & 0.601 \\  
M14 & UBSM-cosJ\&C (this work) & 0.615 & 0.648 & 0.631 & 0.699 & 0.638 & 0.667 & 0.709 & 0.646 & 0.676 & 0.658 \\ 
M15 & UBSM-coswJ\&C (this work) & 0.730 & 0.769 & 0.749 & 0.697 & 0.625 & 0.659 & 0.713 & 0.673 & 0.693 & 0.700 \\
M16 & UBSM-Cai & 0.545 & 0.579 & 0.562 & 0.686 & 0.628 & 0.656 & 0.642 & 0.576 & 0.607 & 0.608 \\ 
M17 & \makecell[l]{COM} & 0.793 & 0.809 & 0.801 & 0.773 & 0.708 & 0.739 & 0.789 & 0.783 & 0.786 & \textcolor{blue}{0.776} \\ 
\hline
M18 & Flair & 0.628 & 0.625 & 0.626 & -0.014 & -0.035 & -0.020 & 0.652 & 0.719 & 0.684 & 0.430 \\
M19 & Pyysalo et al. \cite{Pyysalo2013-jy} & 0.713 & 0.706 & 0.709 & 0.754 & 0.641 & 0.693 & 0.744 & 0.803 & 0.773 & 0.725 \\ 
M20 & BioConceptVec$_{word2vec\_sg}$ & 0.742 & 0.743 & 0.742 & 0.751 & 0.652 & 0.698 & 0.738 & 0.800 & 0.768 & 0.736 \\ 
M21 & BioConceptVec$_{word2vec\_cbow}$ & 0.670 & 0.655 & 0.662 & 0.746 & 0.650 & 0.695 & 0.659 & 0.714 & 0.685 & 0.681 \\ 
M22 & Newman-Griffis$_{word2vec\_sgns}$ & 0.771 & 0.763 & 0.767 & 0.764 & 0.641 & 0.697 & \textbf{0.799} & \textbf{0.835} & \textbf{0.817} & 0.760 \\
M23 & Newman-Griffis$_{word2vec\_cbow}$ & 0.675 & 0.686 & 0.681 & 0.746 & 0.647 & 0.693 & 0.697 & 0.768 & 0.731 & 0.701 \\
M24 & Newman-Griffis$_{glove}$ & 0.671 & 0.678 & 0.674 & 0.740 & 0.643 & 0.688 & 0.732 & 0.729 & 0.731 & 0.698 \\
M25 & BioConceptVec$_{glove}$ & 0.547 & 0.585 & 0.565 & 0.720 & 0.648 & 0.682 & 0.624 & 0.694 & 0.657 & 0.635 \\
M26 & BioWordVec$_{int}$ & \textbf{0.831} & 0.806 & 0.818 & 0.766 & 0.686 & 0.724 & 0.757 & 0.735 & 0.746 & \textcolor{blue}{0.763} \\
M27 & BioWordVec$_{ext}$ & 0.752 & 0.725 & 0.738 & 0.756 & 0.673 & 0.712 & 0.736 & 0.729 & 0.732 & 0.727 \\ 
M28 & BioNLP2016$_{win2}$ & 0.697 & 0.693 & 0.695 & 0.699 & 0.594 & 0.642 & 0.691 & 0.759 & 0.724 & 0.687 \\
M29 & BioNLP2016$_{win30}$  & 0.745 & 0.751 & 0.748 & 0.714 & 0.609 & 0.657 & 0.742 & 0.810 & 0.774 & 0.727 \\
M30 & BioConceptVec$_{fastText}$ & 0.091 & 0.262 & 0.135 & 0.416 & 0.456 & 0.435 & 0.178 & 0.264 & 0.212 & 0.261 \\ 
M31 & USE & 0.666 & 0.669 & 0.668 & 0.679 & 0.606 & 0.640 & 0.663 & 0.684 & 0.674 & 0.660 \\ 
M32 & \makecell[l]{BioSentVec} & 0.797 & 0.767 & 0.782 & 0.763 & 0.638 & 0.695 & 0.791 & 0.821 & 0.806 & 0.761 \\
M33 & FastText-SkGr-BioC (this work) & 0.814 & 0.777 & 0.795 & 0.758 & 0.660 & 0.706 & 0.761 & 0.760 & 0.760 & 0.754 \\ 
\hline
M34 & BioBERT Base 1.0 (+ PubMed) & 0.569 & 0.567 & 0.568 & 0.662 & 0.576 & 0.616 & 0.616 & 0.642 & 0.629 & 0.604 \\ 
M35 & BioBERT Base 1.0 (+ PMC) & 0.664 & 0.663 & 0.664 & 0.674 & 0.581 & 0.624 & 0.601 & 0.647 & 0.623 & 0.637 \\ 
M36 & \makecell[l]{BioBERT Base 1.0$_{(PubMed+PMC)}$} & 0.616 & 0.609 & 0.612 & 0.647 & 0.561 & 0.601 & 0.638 & 0.663 & 0.650 & 0.621 \\ 
M37 & BioBERT Base 1.1 (+ PubMed) & 0.668 & 0.647 & 0.657 & 0.712 & 0.616 & 0.661 & 0.643 & 0.663 & 0.653 & 0.657 \\
M38 & BioBERT Large 1.1 (+ PubMed) & 0.557 & 0.546 & 0.551 & 0.695 & 0.622 & 0.657 & 0.579 & 0.650 & 0.612 & 0.607 \\
M39 & NCBI-BlueBERT Base PubMed & 0.682 & 0.668 & 0.675 & 0.679 & 0.565 & 0.617 & 0.668 & 0.719 & 0.693 & 0.662 \\ 
M40 & NCBI-BlueBERT Large PubMed & 0.688 & 0.712 & 0.700 & 0.636 & 0.588 & 0.611 & 0.609 & 0.674 & 0.640 & 0.650 \\ 
M41 & \makecell[l]{NCBI-BlueBERT Base \\ PubMed + MIMIC-III} & 0.537 & 0.536 & 0.536 & 0.733 & 0.624 & 0.674 & 0.548 & 0.553 & 0.550 & 0.587 \\
M42 & \makecell[l]{NCBI-BlueBERT Large \\ PubMed + MIMIC-III} & 0.560 & 0.578 & 0.569 & 0.675 & 0.628 & 0.651 & 0.487 & 0.504 & 0.496 & 0.572 \\
M43 & SciBERT & 0.653 & 0.616 & 0.634 & 0.727 & 0.643 & 0.683 & 0.604 & 0.682 & 0.641 & 0.652 \\
M44 & ClinicalBERT & 0.415 & 0.483 & 0.447 & 0.652 & 0.566 & 0.606 & 0.470 & 0.500 & 0.485 & 0.512 \\ 
M45 & PubMedBERT (abstracts) & 0.502 & 0.524 & 0.513 & 0.626 & 0.531 & 0.575 & 0.479 & 0.645 & 0.550 & 0.546 \\ 
M46 & \makecell[l]{PubMedBERT \\ (abstracts+full text)} & 0.659 & 0.651 & 0.655 & 0.712 & 0.590 & 0.645 & 0.596 & 0.675 & 0.633 & 0.644 \\ 
M47 & ouBioBERT-Base, Uncased & 0.687 & 0.729 & 0.707 & 0.707 & 0.583 & 0.639 & 0.670 & 0.694 & 0.682 & \textcolor{blue}{0.676} \\ 
M48 & BioClinicalBERT & 0.416 & 0.447 & 0.431 & 0.646 & 0.562 & 0.601 & 0.472 & 0.478 & 0.475 & 0.502 \\ 
M49 & BioDischargesummaryBERT & 0.376 & 0.397 & 0.387 & 0.637 & 0.565 & 0.599 & 0.385 & 0.465 & 0.421 & 0.469 \\
M50 & DischargesummaryBERT & 0.395 & 0.465 & 0.427 & 0.655 & 0.567 & 0.608 & 0.376 & 0.407 & 0.391 & 0.475 \\
\hline
\end{tabular}}{}
\label{tab:table_results}
\end{adjustwidth}
\end{table*}

\begin{table*}[!h]
\begin{adjustwidth}{-2.25in}{0in} 
\caption{Comparison of results for the ``best" and the ``worst" pre-processing configurations for the best-performing methods of each family in table \ref{tab:table_results}. The last column shows the t-Student p-values comparing the best and worst configurations.}
{\small \begin{tabular}{@{}lllcccccccccccccc@{}}
\hline
 & & & \multicolumn{3}{c}{\small{BIOSSES}} & \multicolumn{3}{c}{\small{MedSTS$_{full}$}} & \multicolumn{3}{c}{\small{CTR}} & AVG & \\
ID & Methods & \makecell[l]{Pre-processing \\ configuration} & r & $\rho$ & h & r & $\rho$ & h & r & $\rho$ & h & h & p-val \\
\hline
M4 & \makecell[l]{LiBlock \\ (worst)} & \makecell[l]{ TOK-Whitespace \\ LC-No \\ SW-NLTK2018 \\ CF-None } & 0.779 & 0.793 & 0.786 & 0.736 & 0.676 & 0.704 & 0.765 & 0.717 & 0.741 & 0.744 & \\ 
& & & & & & & & & & & & & 0.000 \\
M4 & \makecell[l]{LiBlock \\ (best)} & \makecell[l]{ TOK-CoreNLP \\ LC-Yes \\ SW-NLTK2018 \\ CF-Default } & 0.820 & 0.828 & 0.824 & 0.769 & 0.710 & 0.739 & 0.793 & 0.808 & 0.800 & 0.788 & \\ 
\hline
M17 & \makecell[l]{COM \\ (worst)} & 
\makecell[l]{ - WBSM-Rada \\ - UBSM-Rada \\ (worst): \\ TOK-Whitespace \\ LC-Yes \\ SW-None \\ CF-None} 
& 0.610 & 0.635 & 0.622 & 0.681 & 0.648 & 0.664 & 0.656 & 0.662 & 0.659 & 0.648 & \\
& & & & & & & & & & & & & 0.000 \\
M17 & \makecell[l]{COM \\ (best)} & \makecell[l]{ - WBSM-Rada \\  - UBSM-Rada \\ (best): \\ TOK-CoreNLP \\ LC-Yes \\ SW-NLTK2018 \\ CF-BIOSSES } & 0.793 & 0.809 & 0.801 & 0.773 & 0.708 & 0.739 & 0.789 & 0.783 & 0.786 & 0.776 & \\
\hline 
M26 & \makecell[l]{BioWordVec$_{int}$ \\ (worst)} & \makecell[l]{ TOK-Whitespace \\ LC-No \\ SW-None \\ CF-None \\ Pooling-Sum} & 0.436 & 0.497 & 0.465 & 0.532 & 0.619 & 0.572 & 0.529 & 0.674 & 0.593 & 0.543 \\
& & & & & & & & & & & & & 0.000 \\
M26 & \makecell[l]{BioWordVec$_{int}$ \\ (best)} & \makecell[l]{ TOK-CoreNLP \\ LC-Yes \\ SW-None \\ CF-BIOSSES \\ Pooling-Min} & 0.831 & 0.809 & 0.820 & 0.764 & 0.682 & 0.721 & 0.761 & 0.736 & 0.748 & 0.763 \\ 
\hline 
M47 & \makecell[l]{OuBioBert \\ (worst)} & \makecell[l]{ TOK- WordPiece \\ LC-Yes \\ SW-BIOSSES \\ CF-Default } & 0.608 & 0.627 & 0.617 & 0.730 & 0.622 & 0.672 & 0.669 & 0.696 & 0.682 & 0.657 & \\
& & & & & & & & & & & & & 0.000 \\
M47 & \makecell[l]{OuBioBert \\ (best)} & \makecell[l]{ TOK-WordPiece  \\ LC-Yes \\ SW-None \\ CF-Default  } & 0.687 & 0.729 & 0.707 & 0.707 & 0.583 & 0.639 & 0.670 & 0.694 & 0.682 & 0.676 & \\
\hline
\end{tabular}}{}
\label{tab:table_comparison_best_worst_prepro}
\end{adjustwidth}
\end{table*}

\begin{table*}[!ht]
\begin{adjustwidth}{-2.25in}{0in} 
\caption{Pearson (r), Spearman ($\rho$) and harmonic ($h$) values obtained in our experiments from the evaluation of ontology similarity methods detailed below in the MedSTS$_{full}$ \cite{Wang2018-oj} dataset for each NER tool.}
\centering
\captionsetup{width=\textwidth}
{\small \begin{tabular}{@{}llccccccccccc@{}}
\hline
& & \multicolumn{3}{c}{\small{MetaMap}} & \multicolumn{3}{c}{\small{MetaMap Lite}} & \multicolumn{3}{c}{\small{cTAKES}} \\
ID & Methods & r & $\rho$ & h  & r & $\rho$ & h  & r & $\rho$ & h \\
\hline
M12 & UBSM-Rada & 0.711 & 0.653 & 0.681 & 0.753 & 0.689 & 0.720 & \textbf{0.764} & \textbf{0.7} & \textbf{0.73} \\
M13 & UBSM-J\&C & 0.576 & 0.547 & 0.561 & \textbf{0.683} & \textbf{0.621} & \textbf{0.65} & 0.634 & 0.549 & 0.588 \\
M14 & UBSM-cosJ\&C & 0.637 & 0.575 & 0.605 & \textbf{0.699} & \textbf{0.638} & \textbf{0.667} & 0.659 & 0.581 & 0.617 \\
M15 & UBSM-coswJ\&C & 0.675 & 0.608 & 0.64 & \textbf{0.722} & \textbf{0.659} & \textbf{0.689} & 0.697 & 0.625 & 0.659 \\
M16 & UBSM-Cai & 0.606 & 0.555 & 0.58 & \textbf{0.686} & \textbf{0.628} & \textbf{0.656} & 0.635 & 0.552 & 0.591 \\
M17 & COM & 0.758 & 0.692 & 0.724 & 0.770 & 0.706 & 0.737 & \textbf{0.773} & \textbf{0.708} & \textbf{0.739} \\
\hline
\end{tabular}}{}
\label{tab:ner_comparison}
\end{adjustwidth}
\end{table*}

\begin{table*}[!ht]
\begin{adjustwidth}{-2.25in}{0in}
\caption{Harmonic score obtained by each combination of a sentence similarity method with a NER tool in the evaluation of the three sentence similarity datasets. The p-values shown in this table are obtained by using the method for building uniform size datasets detailed above. The last column shows the p-values corresponding to the t-Student test comparing the performance of each combination with the best pair in each group.}
\centering
\begin{tabular}{lllccccc}
\hline
ID & Method & NER tool  & \makecell[c]{BIOSSES \\ $h$} & \makecell[c]{MedSTS \\ $h$} & \makecell[c]{CTR \\ $h$} & \makecell[c]{Avg \\ $h$} & p-value \\
\hline
& & cTAKES & 0.800 & 0.730 & 0.785 & 0.772 & --- \\
M12 & UBSM-Rada & MetamapLite & 0.744 & 0.72 & 0.785 & 0.751 & 0.011 \\
& & Metamap & 0.742 & 0.680 & 0.723 & 0.715 & 0.000 \\
\hline
& & MetamapLite & 0.55 & 0.65 & 0.602 & 0.601 & --- \\
M13 & UBSM-J\&C & cTAKES & 0.595 & 0.588 & 0.552 & 0.578 & 0.000 \\
& & Metamap & 0.316 & 0.561 & 0.234 & 0.37 & 0.000 \\
\hline
& & MetamapLite & 0.631 & 0.667 & 0.674 & 0.657 & --- \\
M14 & UBSM-cosJ\&C & cTAKES & 0.681 & 0.617 & 0.626 & 0.641 & 0.002 \\
& & Metamap & 0.537 & 0.605 & 0.434 & 0.525 & 0.000 \\
\hline
& & cTAKES & 0.749 & 0.659 & 0.693 & 0.700 & --- \\
M15 & UBSM-coswJ\&C & MetamapLite & 0.678 & 0.689 & 0.732 & 0.700 & 0.018 \\
& & Metamap & 0.656 & 0.64 & 0.551 & 0.616 & 0.005 \\
\hline
& & MetamapLite & 0.562 & 0.656 & 0.607 & 0.608 & --- \\
M16 & UBSM-Cai & cTAKES & 0.616 & 0.591 & 0.571 & 0.593 & 0.001 \\
& & Metamap & 0.419 & 0.58 & 0.318 & 0.439 & 0.000 \\
\hline
& & cTAKES & \textbf{0.801} & \textbf{0.739} & 0.786 & \textbf{0.776} & --- \\
M17 & COM & MetamapLite & 0.788 & 0.737 & \textbf{0.789} & 0.772 & 0.052 \\
& & Metamap & 0.792 & 0.724 & 0.768 & 0.761 & 0.004 \\
\hline
\end{tabular}
\label{tab:pvalues_ner}
\end{adjustwidth}
\end{table*}

\begin{table*}[!ht]
\begin{adjustwidth}{-2.25in}{0in} 
\caption{Pearson (r) and  Spearman ($\rho$) correlation values, harmonic score ($h$), and harmonic average (AVG) score obtained by the LiBlock method in combination with each NER tool using the best pre-processing configuration detailed in \ref{tab:table_pre-processing_methods_selected}. In addition, last column (p-val) report the p-values for the comparison of the LiBlock method with cTAKES and the remaining NER combinations. }
{\small \begin{tabular}{@{}llcccccccccccccc@{}}
\hline
 & & \multicolumn{3}{c}{\small{BIOSSES} \cite{Sogancioglu2017-rc}} & \multicolumn{3}{c}{\small{MedSTS$_{full}$} \cite{Wang2018-oj}} & \multicolumn{3}{c}{\small{CTR} \cite{Lithgow-Serrano2019-si}} & AVG & \\
ID & \makecell[c]{Sentence similarity \\ methods} & r & $\rho$ & h & r & $\rho$ & h & r & $\rho$ & h & h & p-val\\
\hline
M4 & LiBlock-cTAKES & \textbf{0.820} & \textbf{0.828} & \textbf{0.824} & 0.769 & \textbf{0.710} & \textbf{0.739} & 0.793 & \textbf{0.808} & 0.800 & \textbf{0.788} & -\\ 
M4 & LiBlock-noNER & 0.814 & 0.823 & 0.819 & \textbf{0.770} & 0.709 & 0.738 & \textbf{0.795} & 0.805 & 0.800 & 0.786 & 0.14\\ 
M4 & LiBlock-MetamapLite & 0.799 & 0.819 & 0.809 & 0.763 & 0.705 & 0.733 & 0.794 & \textbf{0.808} & \textbf{0.801} & 0.781 & 0.015 \\ 
M4 & LiBlock-Metamap & 0.807 & 0.826 & 0.816 & 0.753 & 0.690 & 0.720 & 0.792 & 0.807 & 0.799 & 0.779 & 0.003 \\
\hline
\end{tabular}}{}
\label{tab:table_results_liblock}
\end{adjustwidth}
\end{table*}

\begin{table*}[!ht]
\begin{adjustwidth}{-2.25in}{0in}
\caption{Raw and pre-processes sentence pairs obtaining the lowest and highest similarity error $E_{sim}$ together with their corresponding Normalized human similarity score (Human) and normalized similarity value (Method) estimated by the LiBlock (M4) method for the raw and pre-processed sentence pairs with the lowest (L) and highest (H) similarity error $E_{sim}$.}
\centering
\small
\begin{tabular}{cllcc}
\hline
$E_{sim}$ & Input sentence & \makecell[l]{Pre-processed sentence \\ analyzed by the method} & \makecell[l]{Human} & \makecell[l]{Method} \\
\hline
L
 & \makecell[l]{s1: ``Centrosomes increase both in \\ size and in microtubule-nucleating \\ capacity just before mitotic entry." \\ \\ s2: ``Functional studies showed that, when \\ introduced into cell lines, miR-146a was \\ found to promote cell proliferation  in cervical \\ cancer cells, which suggests that miR-146a works \\ as an oncogenic miRNA in these cancers."}
 & \makecell[l]{s1: ``C0242608 increase size C0026046 \\ nucleating  capacity mitotic entry" \\ \\ s2: ``functional studies showed introduced \\ C0007634 lines mir 146a found promote \\ C0007634 C0334094 C4048328 C0007634 \\ suggests mir 146a  works oncogenic \\ mirna C0006826"} 
 & \multirow{5}{*}{0.0} 
 & \multirow{5}{*}{0.0} \\
 \hdashline
 H
 & \makecell[l]{s1: ``Consequently miRNAs have been \\ demonstrated to act either as \\ oncogenes (e.g., miR-155, miR-17−5p \\ and miR-21) or tumor suppressors (e.g., \\ miR-34, miR-15a, miR-16−1 and let-7)" \\ \\ s2: ``Given the extensive involvement of \\ miRNA in physiology, dysregulation of  \\ miRNA expression can be associated with \\ cancer pathobiology including oncogenesis], \\ proliferation, epithelial-mesenchymal \\ transition, metastasis, aberrations in \\ metabolism,  and angiogenesis, among others"}
 & \makecell[l]{s1: ``consequently mirnas demonstrated \\ C0427611 either oncogenes e g mir 155 mir \\ 17 5p mir 21 C0027651 suppressors \\  e g mir 34 mir 15a mir 16 1 let 7" \\ \\ s2: ``given extensive involvement mirna  \\ physiology dysregulation mirna C0185117 \\ associated C0006826 pathobiology including \\ oncogenesis C0334094  epithelial mesenchymal \\ transition metastasis aberrations C0025519 \\ angiogenesis  among others"} 
 & \multirow{5}{*}{0.7} 
 & \multirow{5}{*}{0.0} \\
   \hline
\end{tabular}
\label{tab:raw_best_worst_sentences_string}
\end{adjustwidth}
\end{table*}

\begin{table*}[!h]
\begin{adjustwidth}{-2.25in}{0in}
\caption{Raw and pre-processes sentence pairs obtaining the lowest and highest similarity error $E_{sim}$ together with their corresponding Normalized human similarity score (Human) and normalized similarity value (Method) estimated by the COM (M17) method for the raw and pre-processed sentence pairs with the lowest (L) and highest (H) similarity error $E_{sim}$. We show the raw and pre-processed sentence pairs evaluated by the WBSM and UBSM similarity methods that make up the COM method. The UBSM method use the cTAKES NER tool.}
\centering
\small
\begin{tabular}{cllcc}
\hline
$E_{sim}$ & Input sentence & \makecell[l]{Pre-processed sentence \\ analyzed by the method} & \makecell[l]{Human} & \makecell[l]{Method} \\
\hline
Low
 & \makecell[l]{s1: ``The in vivo data is still preliminary \\  and other potential roadblocks such as \\ drug  resistance have not been examined." \\ \\ s2: ``The GEM model used in this study \\ retains wild-type Tp53, suggesting \\ that the tumors successfully treated \\ with bortezomib and fasudil might \\ not be as aggressive as those \\ in most NSCLC patients"}
 & \makecell[l]{s1, WBSM-Rada: ``vivo data still preliminary \\ potential roadblocks drug resistance examined" \\ s1, UBSM-Rada: ``vivo data still preliminary potential \\ roadblocks C0013227 resistance examined" \\ \\ s2, WBSM-Rada: ``gem model used study retains \\ wild type tp53 suggesting tumors successfully treated \\ bortezomib fasudil might aggressive nsclc patients" \\ s2, UBSM-Rada: ``gem model used study retains wild \\ type tp53 suggesting C0027651 successfully treated \\ C1176309 fasudil might aggressive C0007131 patients"} 
 & \multirow{5}{*}{0.0} 
 & \multirow{5}{*}{0.0} \\ 
  \hdashline
 High
 & \makecell[l]{s1: ``The oncogenic activity of mutant Kras \\ appears dependent on functional Craf, \\ but not on Braf" \\ \\ s2: ``Notably, c-Raf has recently been \\ found essential for development \\ of K-Ras-driven NSCLCs"}
 & \makecell[l]{s1, WBSM-Rada: ``oncogenic activity mutant kras \\ appears dependent functional craf braf" \\ s1, UBSM-Rada: ``oncogenic C0026606 mutant \\ kras appears dependent functional craf braf" \\ \\ s2, WBSM-Rada: ``notably c raf recently found \\ essential development k ras driven nsclcs" \\ s2, UBSM-Rada: ``notably c raf recently \\ found essential development k C0525678 \\ driven nsclcs"} 
 & \multirow{5}{*}{0.75} 
 & \multirow{5}{*}{0.0} \\
   \hline
\end{tabular}
\label{tab:raw_best_worst_sentences_ont}
\end{adjustwidth}
\end{table*}

\begin{table*}[!h]
\begin{adjustwidth}{-2.25in}{0in}
\caption{Raw and pre-processes sentence pairs obtaining the lowest and highest similarity error $E_{sim}$ together with their corresponding Normalized human similarity score (Human) and normalized similarity value (Method) estimated by the BioWordVec$_{int}$ (M26) method for the raw and pre-processed sentence pairs with the lowest (L) and highest (H) similarity error $E_{sim}$.}
\centering
\small
\begin{tabular}{cllcc}
\hline
$E_{sim}$ & Input sentence & \makecell[l]{Pre-processed sentence \\ analyzed by the method} & \makecell[l]{Human} & \makecell[l]{Method} \\
\hline
Low
 & \makecell[l]{s1: ``The up-regulation of miR-146a \\ was also detected in cervical \\ cancer  tissues." \\ \\ s2: ``The expression of miR-146a \\ has been found to be up-regulated \\ in cervical cancer."}
 & \makecell[l]{s1: ``the up regulation of mir 146a \\ was also detected in cervical \\ cancer tissues" \\ \\ s2: ``the expression of mir 146a \\ has been found to be up regulated in  \\ cervical cancer"} 
 & \multirow{5}{*}{1.0} 
 & \multirow{5}{*}{0.986} \\ 
  \hdashline
 High
 & \makecell[l]{s1: ``This oxidative branch activity \\ is elevated in comparison to many \\ cancer cell lines, where the \\ oxidative branch is typically reduced \\ and accounts for <20\% of the carbon \\ flow through PPP." \\ \\ s2: ``The Downward laboratory went \\ all the way from identifying \\ GATA2 as a novel synthetic lethal gene \\ to validating it using \\ Kras-driven GEM models."}
 & \makecell[l]{s1: ``this oxidative branch activity \\ is elevated in comparison to many \\ cancer cell lines where the \\ oxidative branch is typically reduced \\ and accounts for < 20 \% of the \\ carbon flow through ppp" \\ \\ s2: ``the downward laboratory went \\ all the way from identifying gata2 \\ as a novel synthetic lethal gene \\ to validating it using kras driven \\ gem models"} 
 & \multirow{5}{*}{0.0} 
 & \multirow{5}{*}{0.912} \\
   \hline
\end{tabular}
\label{tab:raw_best_worst_sentences_embedd}
\end{adjustwidth}
\end{table*}

\begin{table*}[!h]
\begin{adjustwidth}{-2.25in}{0in}
\caption{Raw and pre-processes sentence pairs obtaining the lowest and highest similarity error $E_{sim}$ together with their corresponding Normalized human similarity score (Human) and normalized similarity value (Method) estimated by the OuBioBert (M47) method for the raw and pre-processed sentence pairs with the lowest (L) and highest (H) similarity error $E_{sim}$.}
\centering
\small
\begin{tabular}{cllcc}
\hline
$E_{sim}$ & Input sentence & \makecell[l]{Pre-processed sentence \\ analyzed by the method} & \makecell[l]{Human} & \makecell[l]{Method} \\
\hline
Low
 & \makecell[l]{s1: ``Expression of an activated form \\ of Ras proteins can induce senescence in  \\ some primary fibroblasts." \\ \\ s2: ``The senescent state has been \\ observed to be inducible in certain  \\ cultured cells in response to high \\ level expression of genes  \\ activated such as the ras oncogene."}
 & \makecell[l]{s1: ``expression activated form ras proteins \\ induce senescence primary fibroblasts" \\ \\ s2: ``senescent state observed inducible \\ certain cultured cells response  high level \\ expression genes activated ras oncogene"} 
 & \multirow{5}{*}{0.9} 
 & \multirow{5}{*}{0.908} \\ 
  \hdashline
 High
 & \makecell[l]{s1: ``The in vivo data is still preliminary \\ and other potential roadblocks such as drug \\ resistance have not been examined." \\ \\ s2: ``The GEM model used in this study \\ retains wild-type Tp53, suggesting \\ that the tumors successfully treated with \\ bortezomib and fasudil might not be as \\ aggressive as those in most NSCLC patients"}
 & \makecell[l]{s1: ``vivo data still preliminary potential \\ road bl ocks drug resistance examined" \\ \\ s2: ``gem model used study retains wild \\ type tp53 suggesting tumors successfully \\ treated bortezomib fas udi l might \\ aggressive nsclc patients"} 
 & \multirow{5}{*}{0.0} 
 & \multirow{5}{*}{0.773} \\
   \hline
\end{tabular}
\label{tab:raw_best_worst_sentences_bert}
\end{adjustwidth}
\end{table*}

\clearpage

\begin{figure}[!h]
\centering
\caption{Probability Density Function (PDF) and mean value of the similarity error ($E_{sim}$) obtained by the best-performing methods in the evaluation of each dataset as follows: (a) BIOSSES, (b) MedSTS, and (c) CTR.}
\label{fig:probabilityerrordistribution}
\end{figure}

\begin{table*}[!h]
\caption{Comparison of the mean, minimum and maximum similarity scores of the Normalized Human similarity scores (Human) and the estimated valued returned by the best-performing methods of each family in the evaluation of the three biomedical datasets.}
    \centering
    \begin{tabular}{clccc}
    
    \\
    \multicolumn{5}{l}{BIOSSES dataset} \\
    \hline
    ID & Method & \makecell[l]{Mean \\ similarity} & \makecell[l]{Minimum \\ similarity} & \makecell[l]{Maximum \\ similarity} \\
    \hline
    - & Human & 0.549  & 0 & 1  \\
      M4   & LiBlock (this work) & 0.194 & 0 & 0.506 \\
      M17   & COM \cite{Sogancioglu2017-rc} & 0.22 & 0 & 0.596 \\
      M26  & BioWordVec$_{int}$ \cite{Zhang2019-qq} & 0.933 & 0.858 & 0.987  \\
      M47  & OuBioBert \cite{Wada2020-nw} & 0.808  & 0.582 & 0.936  \\
      
      \\
      \multicolumn{5}{l}{MedSTS dataset} \\
      \hline
    ID & Method & \makecell[l]{Mean \\ similarity} & \makecell[l]{Minimum \\ similarity} & \makecell[l]{Maximum \\ similarity} \\
    \hline
    - & Human  & 0.632 & 0 & 1 \\
      M4   & LiBlock (this work) & 0.611  & 0 & 1 \\
      M17   & COM \cite{Sogancioglu2017-rc} & 0.631  & 0 & 1 \\
      M26  & BioWordVec$_{int}$ \cite{Zhang2019-qq} & 0.957  & 0.832 &  1 \\
      M47  & OuBioBert \cite{Wada2020-nw} & 0.885  & 0.437 &  0.997 \\
      
      \\
      \multicolumn{5}{l}{CTR dataset} \\
      \hline
    ID & Method & \makecell[l]{Mean \\ similarity} & \makecell[l]{Minimum \\ similarity} & \makecell[l]{Maximum \\ similarity} \\
    \hline
    - & Human & 0.254  & 0 &  1 \\
      M4   & LiBlock (this work) & 0.103  & 0 & 0.743 \\
      M17   & COM \cite{Sogancioglu2017-rc} & 0.118  & 0 & 0.793 \\
      M26  & BioWordVec$_{int}$ \cite{Zhang2019-qq} & 0.898  & 0.752 &  0.992 \\
      M47  & OuBioBert \cite{Wada2020-nw} & 0.724  & 0.472 &  0.98 \\
    \end{tabular}
    \label{tab:statsmethods}
\end{table*}

\section*{Discussion}
\label{sec:discussion}

\subsection*{Comparison of string-based methods}

\emph{LiBlock (M4) obtains the highest average harmonic score among the family of string-based methods and significantly outperforms all of them.} This conclusion can be drawn by looking at the average column in table \ref{tab:table_results} for this group of methods and checking the p-values reported in table A.1, such as Block Distance (p-value=0.000), Jaccard (p-value=0.000), QGram (p-value=0.000), Overlap Coefficient (p-value=0.000), and Levenshtein (p-value=0.000).

\emph{LiBlock (M4) obtains the highest Pearson correlation value in the BIOSSES and MedSTS datasets among the family of string-based methods, whilst Block Distance (M3) obtains the highest Pearson correlation in the CTR dataset}. This conclusion can be drawn by looking the results for the first group of methods detailed in table \ref{tab:table_results}.

\emph{LiBlock (M4) obtains the highest Spearman correlation value in all datasets among the family of string-based methods.} This conclusion can be drawn by looking at the results for the first group of methods detailed in table \ref{tab:table_results}.

\emph{LiBlock (M4) obtains the highest harmonic score in all datasets among the family of string-based methods}. This conclusion can be drawn by looking the results for the first group of methods detailed in table \ref{tab:table_results}.

\subsection*{Comparison of Ontology-based methods}

\emph{COM (M17) obtains the highest average harmonic score among the family of ontology-based methods significantly outperform all of them, with the only exception of WBSM-Rada (M7)}. This conclusion can be drawn by looking at the average column in table \ref{tab:table_results} for the second group of methods and checking the p-value shown in table A.1 for the comparison of COM (M17) with WBSM-Rada (M7) (p-value=0.088).

\emph{COM (M17) obtains the highest Pearson correlation value in the BIOSSES and CTR datasets among the family of ontology-based methods, whilst the WBSM-Rada (M7) methods obtain the highest Pearson correlation value in the MedSTS dataset.} This conclusion can be drawn by looking at the second group of methods in \ref{tab:table_results}.

\emph{COM (M17) obtains the highest Spearman correlation values in the BIOSSES dataset among the family of ontology-based methods, whilst WBSM-Rada (M7) and UBSM-Rada (M12) do it in the MedSTS and CTR datasets, respectively}. This conclusion can be drawn by looking at the second group of methods in \ref{tab:table_results}.

\emph{COM (M17) obtains the highest harmonic score in the BIOSSES and CTR datasets among the family of ontology-based methods, whilst WBSM-Rada (M7) does it in the MedSTS dataset.} This conclusion can be drawn by looking at the second group of methods detailed in table \ref{tab:table_results}.

\subsection*{Comparison of embeddings methods}

\emph{BioWordVec$_{int}$ (M26) obtains the highest average harmonic score in all datasets among the family of embedding methods detailed in table \ref{tab:descriptionmethods_sentenceEmbeddings}, and significantly outperforms all of them.} This conclusion can be drawn by looking at the third group of methods in table \ref{tab:table_results} and checking the p-values reported in table A.1, which compare the harmonic score values obtained by the BioWordVec$_{int}$ (M26) method with the rest of methods from the same family, such as FastText-SkGr-BioC (p-value=0.032), BioWordVec$_{ext}$ (p-value = 0.007), and BioSentVec (p-value=0.022) among others.

\emph{BioWordVec$_{int}$ (M26) obtains the highest Pearson correlation value in the BIOSSES and MedSTS datasets among the family of embedding methods, whilst the Newman\linebreak-Griffis$_{word2vec\_sgns}$ (M22) model does it in the CTR dataset}. This  conclusion can be drawn by looking the results for third group of methods detailed in table \ref{tab:table_results}.

\emph{BioWordVec$_{int}$ (M26) obtains the highest Spearman correlation in the BIOSSES and MedSTS datasets among the family of embedding methods, whilst the Newman-Griffis$_{word2vec\_sgns}$ (M22) model does it in the CTR dataset.} This later conclusion can be drawn by looking the results for the third group of measures detailed in table \ref{tab:table_results}.

\emph{BioWordVec$_{int}$ (M26) obtains the highest harmonic score in the BIOSSES and MedSTS datasets among the family of embedding methods, whilst the Newman-Griffis$_{word2vec\_sgns}$ (M22) model does it in the CTR dataset.} This later conclusion can be drawn by looking the results for the third group of measures detailed in table \ref{tab:table_results}.

\subsection*{Comparison of BERT-based methods}

\emph{OuBioBERT (M47) obtains the highest average harmonic score among the family of BERT-based methods. However, it does not significantly outperform all of them.} This conclusion can be drawn by looking at the last group of methods in table \ref{tab:table_results} and checking the p-values reported in table A.1. Table A.1 shows that  ouBioBERT obtains p-values higher than 0.05 when it is compared with many BERT-based methods, such as BioBERT Large 1.1 (p-value=0.224) and PubMedBERT (abstracts+full text) (p-value=0.101) among others.

\emph{NCBI-BlueBERT Large PubMed (M40) obtains the highest Pearson correlation value in the BIOSSES dataset among the family of BERT-based methods, whilst the NCBI-BlueBERT Base PubMed + MIMIC-III (M41) and the ouBioBERT (M47) models do it in the MedSTS and the CTR datasets, respectively.} This later conclusion can be drawn by looking at the last group of measures detailed in table \ref{tab:table_results}.

\emph{ouBioBERT (M47) obtains the highest Spearman correlation value in the BIOSSES dataset among the family of BERT-based methods, whilst SciBERT (M43) and NCBI-BlueBERT Base PubMed (M39) do it in the MedSTS and CTR datasets, respectively.} This conclusions can be drawn by looking at the last group of measures detailed in table \ref{tab:table_results}.

\emph{ouBioBERT (M47) obtains the highest harmonic score in the BIOSSES dataset among the family of BERT-based methods, whilst SciBERT (M43) and NCBI-BlueBERT Base PubMed (M39) do it in the MedSTS and CTR datasets, respectively.} This conclusion can be drawn by looking at the last group of measures detailed in table \ref{tab:table_results}.

\subsection*{Comparison of all methods}

\emph{LiBlock (M4) obtains the highest average harmonic score for all the methods evaluated herein, and significantly outperforms all the methods based on embeddings and language models. However, there is no a statistically significant difference in performance with the ontology-based methods COM (M17) and WBSM-Rada (M7).} This conclusion can be drawn by looking at the average column in table \ref{tab:table_results} and checking the p-value reported in table A.1, which compare the harmonic score obtained by the LiBlock method with the COM (p-value=0.121) and WBSM-Rada (p-value=0.098) methods.

\emph{BioWordVec$_{int}$ (M26) obtain the highest Pearson correlation values in the BIOSSES dataset among all methods evaluated herein, whilst WBSM-Rada (M7) and Newman-Griffis$_{word2vec\_sgns}$ (M22) do it in the MedSTS and CTR datasets, respectively.} This conclusion can be drawn by looking at the bold values detailed in table \ref{tab:table_results}.

\emph{LiBlock (M4) obtains the highest Spearman correlation value in the BIOSSES and MedSTS datasets among all methods evaluated herein, whilst Newman-Griffis$_{word2vec\_sgns}$ (M22) does it in the CTR dataset.} This conclusions can be drawn by looking at the bold values detailed in table \ref{tab:table_results}.

\emph{LiBlock (M4) obtains the highest harmonic score in the BIOSSES dataset among all methods evaluated herein, whilst WBSM-Rada (M7) and Newman-Griffis$_{word2vec\_sgns}$ (M22) do it in the MedSTS and CTR datasets, respectively.} This conclusion can be drawn by looking at the bold values detailed in table \ref{tab:table_results}.

\emph{COM (M17) obtains the second highest average harmonic score among all methods evaluated herein, and it is able to outperform significantly all methods with the only exception of LiBlock (M4) and WBSM-Rada (M7).} This conclusion can be drawn by looking at the bold values detailed in table \ref{tab:table_results} and checking the p-value reported in table A.1.

\subsection*{Non ML-based methods versus ML-based ones}

\emph{The string-based methods LiBlock (M4) and Block Distance (M3) obtain a higher average harmonic score than all the embedding-based methods in all datasets. Moreover, the string-based method LiBlock (M4) significantly outperforms all the methods based on embedding models.} This conclusion can be drawn by looking at the average column in table \ref{tab:table_results} and checking the p-values reported in table A.1, such as BioWordVec$_{int}$ (p-value 0.003), FastText-SkGr-BioC (p-value 0.002), BioConceptVec$_{glove}$ (p-value 0.001), Flair (p-value 0.027), and the rest of embedding-based methods (p-value 0.000). 

\emph{All string-based methods obtain a higher average harmonic score than all the BERT-based methods considering all datasets, with the only exception of the Levenshtein distance (M5). Moreover, most string-based methods significantly outperforms all BERT-based methods, with the only exception of the Levenshtein distance (M5).} This conclusion can be drawn by looking at the average column in table \ref{tab:table_results} and checking the p-values reported in table A.1.

\emph{The ontology-based methods COM (M17), WBSM-Rada (M7) and UBSM-Rada (M12) obtain a higher average harmonic score than all the embedding-based methods considering all datasets and significantly outperforms all of them.} This conclusion can be drawn by looking at the average column in table \ref{tab:table_results} and checking the p-values reported in table A.1, which compare the harmonic scores obtained by COM (M17), WBSM-Rada (M7) and UBSM-Rada (M12) with all the embedding-based methods.

\emph{The ontology-based methods UBSM-Rada (M12), WBSM-Rada (M7), COM (M17) and UBSM-coswJ\&C (M15) obtain a higher average harmonic score than all the BERT-based methods. Moreover, the ontology-based methods UBSM-Rada (M12), WBSM-Rada (M7), and COM (M17) significantly outperforms all the BERT-based methods.} This conclusion can be drawn by looking at the average column in table \ref{tab:table_results} and checking the p-values reported in table A.1.

\emph{All embedding methods obtain a higher average harmonic score than all BERT-based methods, with the only exceptions of Flair (M18), BioConceptVec$_{glove}$ (M25), BioConceptVec$_{fastText}$ (M30) and USE (M31).}  This conclusion can be drawn by looking at the last column in table \ref{tab:table_results}.

\emph{BioWordVec$_{int}$ (M26) obtains a higher average harmonic score than all the BERT-based methods considering all datasets and significantly outperforms all of them.} This conclusion can be drawn by looking at the average column in table \ref{tab:table_results} and checking the p-values reported in table A.1, which compare the harmonic scores obtained by BioWordVec$_{int}$ (M26) with all the BERT-based methods, such as SciBERT (p-value 0.001), NCBI-BlueBERT Base PubMed + MIMIC-III (p-value 0.002), BioBERT Large 1.1 (p-value 0.001), and the rest of BERT-based methods (p-value 0.000).

\subsection*{Impact of the NER tools on the ontology-based methods}

This section analyzes the impact of the NER tools on the performance of the sentence similarity methods, and studies the overall impact of the NER configurations. Table \ref{tab:ner_comparison} shows the results obtained on the performance of NER tools for the sentence similarity methods evaluated in the MedSTS dataset \cite{Wang2018-oj}, whilst table \ref{tab:pvalues_ner} shows the harmonic and average harmonic scores, as well as the resulting p-values comparing the harmonic score of the best-performing NER tool for each ontology-based method in the three datasets with the harmonis scores obtained by the other two NER tools.

\emph{MetamapLite obtains the highest Pearson, Spearman, and harmonic scores for the MedSTS dataset in combination with UBSM-J\&C (M13), UBSM-cosJ\&C (M14), UBSM-coswJ\&C (M15) and UBSM-Cai (M16), whilst cTAKES obtains the highest Pearson, Spearman and harmonic scores for the MedSTS dataset in combination with UBSM-Rada (M12) and COM (M17).} This later conclusion can be drawn by looking at the results shown in table \ref{tab:ner_comparison}.

\emph{cTAKES obtains the highest average harmonic score for the three datasets in combination with UBSM-Rada (M12), UBSM-coswJ\&C (M15) and COM (M17) methods, whilst MetamapLite obtains the highest average harmonic score for the three datasets in combination with UBSM-J\&C (M13), UBSM-cosJ\&C (M14) and UBSM-Cai (M16).} This conclusion can be drawn by looking at the harmonic scores of the NER tools in table \ref{tab:pvalues_ner}.

\emph{cTAKES combined with COM (M17) obtains the best-performing results of ontology-based methods for the three datasets.} This conclusion can be drawn by looking at the average harmonic scores column shown in table \ref{tab:pvalues_ner}.

\emph{cTAKES is the best-performing tool in combination with the UBSM-Rada (M12), UBSM-coswJ\&C (M15), and COM (M17) methods in the three datasets, and significantly outperforms MetamapLite and Metamap or the two former methods. However, there is no a statistically significant diference regarding the Metamap tools when it is combined with the COM (M17) method.} This conclusion can be drawn by looking at the average harmonic scores and p-values shown in table \ref{tab:pvalues_ner}. 

\emph{MetamapLite is the best-performing tool in combination with the UBSM-J\&C (M13),  UBSM-cosJ\&C (M14), and UBSM-Cai (M16) methods in the three datasets, and significantly outperforms cTAKES and Metamap.} This conclusion can be drawn by looking at the average harmonic scores and p-values shown in table \ref{tab:pvalues_ner}. 

\emph{The choice of the best NER tool for each method significantly impact their performance in most cases.} This conclusion follows from the conclusions above.

\paragraph{Answering RQ3.} Our results show that the ontology-based methods obtain their best performance in the task of biomedical sentence similarity when they use either MetamapLite or cTAKES. Thus, Metamap should not be used in combination with any of the ontology-based methods evaluated herein in this later task. Likewise, the results and p-values reported table \ref{tab:pvalues_ner} show that there is a significant difference in the performance of each ontology-based method according to the NER tool used in most cases. The conclusions above confirm that the selection of the NER tool significantly impacts the performance of the sentence similarity methods using it.

\subsection*{Impact of the NER tools on the new LiBlock measure}

This section analyzes the impact of the NER tools on the new $sim_{LiBk}$ similarity measure. Table \ref{tab:table_results_liblock} shows the results obtained by the $sim_{LiBk}$ measure in the three biomedical datasets using its best pre-processing configuration, and annotating the sentences with all the combinations of NER tools. In addition, the aforementioned table details the resulting p-values comparing the best-performing LiBlock-NER combination with the combinations based on the other two NER tools.

\emph{LiBlock-cTAKES obtains the highest average harmonic score for the three datasets among the LiBlock-NER combinations. However, it does not significantly outperform LiBlock with no use of a NER tool.} This conclusion can be drawn by looking at the average column in table \ref{tab:table_results_liblock} and checking the p-values in the last column. This conclusion is especially relevant because it shows that there is no a statistically significant difference between using a NER tool like cTAKES or not using it in the case of the LiBlock measure. We conjecture that this later conclusion could be caused by two reasons, firstly the incapability of LiBlock to capture semantic relationships beyond the synonymy, and secondly the current limitations of cTakes to recognize all mentions of biomedical entities.

\emph{LiBlock-cTAKES obtains the highest Pearson correlation value in the BIOSSES dataset among all LiBlock-NER combinations, whilst LiBlock with no use of a NER tool obtains the highest Pearson correlation value in the MedSTS and CTR datasets, respectively.} This conclusion can be drawn by looking the results detailed in table \ref{tab:table_results_liblock}.

\emph{LiBlock-cTAKES obtains the highest Spearman correlation value in the BIOSSES and MedSTS datasets among the LiBlock-NER combinations, whilst LiBlock-cTAKES and LiBlock-MetamapLite obtain the highest Spearman correlation value in the CTR dataset.} This conclusion can be drawn by looking the results detailed in table \ref{tab:table_results_liblock}.

\emph{LiBlock-cTAKES obtains the highest harmonic correlation value in the BIOSSES and MedSTS datasets among the LiBlock-NER combinations, whilst LiBlock-MetamapLite obtains the highest harmonic correlation value in the CTR dataset.} This conclusion can be drawn by looking the results detailed in table \ref{tab:table_results_liblock}.

\subsection*{Impact of the remaining pre-processing stages}

This section analyzes the impact of each pre-processing step on the performance of the sentence similarity methods, except for the NER tools already analyzed in the previous section. Finally, we study the overall impact of the pre-processing configurations.

\subsubsection*{Impact of tokenization}

\emph{The family of string-based methods obtains its best-performing results either by splitting the sentence from the white spaces between words or using the Stanford CoreNLP tokenizer.} This conclusion can be drawn by looking at the table \ref{tab:table_pre-processing_methods_selected}, which summarizes the pre-processing tables detailed in Appendix B.

\emph{The family of ontology-based methods obtains its best-performing results in combination with the Stanford CoreNLP tokenizer.} This conclusion can be drawn by looking at the table \ref{tab:table_pre-processing_methods_selected}.

\emph{The family of methods based on embeddings obtains its best-performing results in combination with the Stanford CoreNLP tokenizer, with the only exception of Flair (M18).} This conclusion can be drawn by looking at the table \ref{tab:table_pre-processing_methods_selected}.

\emph{None method based on strings, ontologies, or embeddings obtain its best-performing results in combination with the BioCNLPTokenizer.} This conclusion can be drawn by looking at the table \ref{tab:table_pre-processing_methods_selected}. Thus, the BioCNLPTokenizer should not be used in combination with any method in the former families in the task of biomedical sentence similarity. On the other hand, we recall that all BERT-based methods evaluated herein can only be used in combination with the WordPiece Tokenizer \cite{Wu2016-en} based on a subword segmentation algorithm, because it is required by the current BERT implementations.

\emph{All families of methods show a strong preference by a specific tokenizer, with the only exception of the string-based one.} This conclusion can be drawn from previous conclusions that confirm the preference of the methods based on ontologies and embeddings by the CoreNLP tokenizer, and the mandatory use of the WordPiece tokenizer by the family of BERT-based methods.

\subsubsection*{Impact of character filtering}

\emph{The family of string-based methods obtains its best-performing results by using either the BIOSSES char-filtering method or the default method which removes the punctuation marks and special symbols from the sentences, with the only exception of the Levenshtein distance method (M5), which does not remove special characters.} This conclusion can be drawn by looking at the table \ref{tab:table_pre-processing_methods_selected}, which summarizes the pre-processing tables detailed in Appendix B.

\emph{All ontology-based methods obtain their best-performing results in combination with the BIOSSES char-filtering method.} This conclusion can be drawn by looking at the table \ref{tab:table_pre-processing_methods_selected}.

\emph{Most of embeddings methods obtain their best-performing results in combination with the default char filtering method. However, Flair (M18), BioWordVec (M26,M27), and BioSentVec (M32) obtain their best-performing results with the BIOSSES char-filtering method.} This conclusion can be drawn by looking at the table \ref{tab:table_pre-processing_methods_selected}.

\emph{The BERT-based methods do not show a noticeable preference pattern by a specific char filtering method, obtaining their best-performing results with the BIOSSES, Blagec2019, or the default one.} This conclusion can be drawn by looking at the table \ref{tab:table_pre-processing_methods_selected}.

\subsubsection*{Impact of stop-words removal}

\emph{All string-based methods obtain their best-performing results in combination with the NLTK2018 stop-word list, with the only exception of the Levenshtein distance (M5).} This conclusion can be drawn by looking at the table \ref{tab:table_pre-processing_methods_selected}, which summarizes the pre-processing tables detailed in Appendix B.

\emph{All ontology-based methods obtain their best-performing results in combination with the NLTK2018 stop-word list, with the only exception of WBSM-J\&C (M8), WBSM-cosJ\&C (M9), which do not remove stop words.} This conclusion can be drawn by looking at the table \ref{tab:table_pre-processing_methods_selected}.

\emph{The methods based on embeddings do not show a noticeable preference pattern by a specific stop-word list, obtaining their best-performing results by using the stop-word list of BIOSSES, NLTK2018, or none.} This conclusion can be drawn by looking at the table \ref{tab:table_pre-processing_methods_selected}.

\emph{The methods based on language models do not show a noticeable preference pattern by a specific stop-word list, obtaining their best-performing results by using the stop-word list of BIOSSES, NLTK2018, or none.} This conclusion can be drawn by looking at the table \ref{tab:table_pre-processing_methods_selected}.

\emph{The best-performing results for the methods based on strings or ontologies show a noticeable preference by the use of the stop-words list NLTK2018.} This conclusion can be drawn by looking at the table \ref{tab:table_pre-processing_methods_selected}.

\subsubsection*{Impact of lower-casing}

\emph{Only 10 of the 50 methods evaluated in this work obtain their best performance by avoiding converting words to lowercase at the sentence pre-processing stage.} This conclusion can be drawn by looking at the tables \ref{tab:table_pre-processing_methods_selected} and \ref{tab:table_results}, and the pre-processing tables detailed in Appendix B. Moreover, these ten aforementioned methods obtain a low performance in our experiments, with the only exception of the BioNLP2016$_{win30}$ (M29) pre-trained model, which obtains the third best Spearman correlation value in the CTR dataset. Thus, our experiments confirm that the lower-casing normalization of the sentences positively impacts the performance of the methods, and it should be considered as default option in any biomedical sentence similarity task.

We conjecture that lower-casing improves the performance of the families of string-based and ontology-based methods because it improves the exact comparison of words. On the other hand, we also conjecture that the impact of lower-casing the sentences on the families of methods based on embeddings and language models strongly depends on the pre-processing methods used in their training.

\subsubsection*{Overall impact of the pre-processing}

To study the overall impact of the pre-processing stage on the performance of the sentence similarity methods, we selected the configuration reporting the highest (best) and lowest (worst) average harmonic score values for each method, as shown in table \ref{tab:table_comparison_best_worst_prepro}. These configurations were selected from a total of 1081 pre-processing configurations reported in Appendix B. 

\emph{The best-performing methods of each family show a statistically significant difference in performance between their best and worst pre-processing configurations.}  This conclusion can be drawn by looking at the average (AVG) and the p-values in table \ref{tab:table_comparison_best_worst_prepro}. 

\paragraph{Answering RQ4.} Our results and the conclusions above show that the pre-processing configurations significantly impact the performance of the sentence similarity methods, and thus, it should be specifically defined for each method. All families of methods show a strong preference by a specific tokenizer, with the only exception of the string-based one. In addition, the BioCNLPTokenizer does not contribute to the best-performing configuration of any method evaluated herein. The family of string-based methods shows a preference pattern of using either the BIOSSES or default char filtering method, whilst all ontology-based methods use the BIOSSES char filtering method, and most embedding methods use the default char filtering method. However, BERT-based methods do not show a noticeable preference pattern by a specific char filtering method. On the other hand, the families of string and ontology-based methods show a noticeable preference pattern by the use of the NLTK2018 stop-words list, whilst the families of embeddings and BERT-based methods do not show a noticeable pattern. Finally, the experiments confirm that the lower-casing normalization of the sentences positively impacts the performance of the methods, and it should be considered as default option in any biomedical sentence similarity task.

\subsection*{The new state-of-the-art}

We set the new state of the art to answer our RQ1 and RQ2 questions as follows.

LiBlock (M4) measure sets the new state of the art for the sentence similarity task in the biomedical domain (see table \ref{tab:table_results}), being the best overall performing method to tackle this later task. Moreover, LiBlock significantly outperforms all the methods based on embeddings and language models. However, LiBlock cannot significantly outperform the COM (M17) and WBSM-Rada (M7) ontology-based methods (see Appendix A.1). Thus, LiBlock is a convincing but non-definitive winner among the biomedical sentence similarity methods evaluated herein.

COM (M17) method sets the new state of the art among the family of ontology-based methods for biomedical sentence similarity, being the best-performing method in this later task (see table \ref{tab:table_results}). COM significantly outperforms all methods based on embeddings and BERT-based language models, as well as all string-based and ontology-based methods with the only exception of LiBlock (M4) and WBSM-Rada (M7) (see Appendix A.1).

BioWordVec$_{int}$ (M26) sets the new state of the art among the family of methods based on pre-trained  embedding models, being the best-performing method in this later task (see table \ref{tab:table_results}), and significantly outperforming the remaining methods in the same family (see Appendix A.1).

OuBioBERT (M47) sets the new state of the art in among the family of methods based on pre-trained BERT models, being the best-performing method in this later task (see table \ref{tab:table_results}). However, OuBioBERT is unable to outperform significantly all remaining methods from the same family (see Appendix A.1).

Finally, our results show that our new string-based method, called LiBlock (M4), obtains the best overall performing results, despite it does not capture the semantic information of the sentences. This is a very noticeable finding because it contradicts a common belief on the potential outperformance of the ontology-based methods integrating word and concept semantics over the non-semantics methods in this similarity task. A second and very noticeable finding is that our non-semantics and non-ML LiBlock method is able to outperform significantly state-of-the-art methods based on large ML models trained with the most recent and advanced word embeddings \cite{Lastra-Diaz2019-ai} and BERT language models  \cite{Devlin2018-eq} in an unsupervised context. This later finding is very remarkable because LiBlock is easy of implementing, easy of evaluating, very efficient (2635 sentence pairs per second with no use of a NER tool), and it requires neither large text resources nor complex algorithms for its training and evaluation, which is a very clear advantage in the biomedical sentence similarity task.

\paragraph{Answering RQ1 and RQ2.} The string-based method LiBlock (M4) obtains the highest average harmonic score in all datasets, and significantly outperforms the remaining string-based methods, as well as all methods based on embeddings and BERT language models, and all the ontology-based methods with the only exceptions of COM (M17) and WBSM-Rada (M7). In addition, LiBlock obtains the highest Spearman correlation values in the BIOSSES and MedSTS datasets, which contains 100 and 1068 sentence pairs respectively.

\subsection*{Main drawbacks and limitations of current methods}

This section analyzes the behaviour of the best-performing methods in each family of sentence similarity methods to answer our RQ5. The best-performing methods of each family, according to the harmonic average value reported in table \ref{tab:table_results}, are LiBlock (M4), COM (M17), BioWordVec$_{int}$ (M26), and OuBioBERT (M47).

\emph{String and ontology-based methods underestimate in average the human similarity value in the BIOSSES and CTR datasets, whilst their average similarity error is close to 0 in the MedSTS dataset.} This conclusion can be drawn by looking at the average similarity error values and the mean error values shown in figure \ref{fig:probabilityerrordistribution} together with the mean values shown in table \ref{tab:statsmethods}. LiBlock and COM obtain mean error values of -0.021 and -0.001 in MedSTS, as shown in figure \ref{fig:probabilityerrordistribution}.b. On the other hand, both methods report a mean similarity score much lower than the mean of the Human normalized score in the BIOSSES and CTR datasets and a mean similarity score close to the Human normalized score in the MedSTS dataset, as shown in table  \ref{tab:statsmethods}.

\emph{The methods based on embeddings and language models overestimate in average the human similarity value in the three datasets.} This conclusion can be drawn by looking at the average similarity error values and the mean error values shown in figure \ref{fig:probabilityerrordistribution}, together with  the mean similarity values shown in table \ref{tab:statsmethods}. The two aforementioned families of methods report a mean similarity score much higher than the mean of the Human normalized score in the three datasets, as show in table \ref{tab:statsmethods}.

\emph{String and ontology-based methods share a similar underestimation behavior, in opposition to the overestimation behaviour shown by the methods based on embeddings and language models, which is very noticeable in the three datasets.} This conclusion can be drawn by looking at the minimum and maximum similarity values columns in table \ref{tab:statsmethods}, and the plots of the probability error distribution function for the three datasets in figure \ref{fig:probabilityerrordistribution}. For instance, despite the human similarity scores are in the range of 0 to 1 n the BIOSSES dataset, as shown in table \ref{tab:statsmethods}, the string and ontology-based methods report similarity scores in the range of 0 to 0.596, whilst the methods based on embeddings and language models report similarity scores in the range of 0.582 to 0.987.

\emph{String and ontology-based methods tend to obtain their best results in sentences with a Human normalized score close to 0, whilst the methods based on embeddings and language models obtain their best results in sentences with a Human normalized score close to 1.} This conclusion can be drawn by looking at the tables \ref{tab:raw_best_worst_sentences_string}, \ref{tab:raw_best_worst_sentences_ont}, \ref{tab:raw_best_worst_sentences_embedd} and \ref{tab:raw_best_worst_sentences_bert}. On the other hand, string and ontology-based methods tend to obtain their worst results in sentences with a Human normalized score close to 1, whilst the methods based on embeddings and language models obtain their worst results in sentences with a Human normalized score close to 0.

\emph{None of the methods for semantic similarity of sentences in the biomedical domain evaluated herein use an explicit syntactic analysis or syntax information to obtain the similarity value.} We conjecture that syntactic analysis would improve the performance in some cases. For instance, the sentences $s1$ and $s2$ with highest $E_{sim}$ in table \ref{tab:raw_best_worst_sentences_string} shows an implicit relation between the concepts "miRNA" and "oncogenesis", which should increase the final semantic similarity score of the sentences. However, none of the methods evaluated herein consider and reward these semantics relationships because its recognition demands some form of syntactic analysis. On the one hand, string and ontology-based methods consider the concepts in a sentence as bags of words, whilst on the other hand the methods based on embeddings and language models implicitly consider the structure of the sentences but not the relationships between the parts of the sentences that are related. 

\emph{Our results show that the family of string-based methods is rewarded by the high frequency of overlapping words in the sentences of the current biomedical datasets, whilst the former methods are not able to deal properly with sentences that are semantically different but not exhibit a word overlapping pattern.} The main advantages of the string-based methods are as follows: (1) they are able to obtain high correlation values without the need of using external resources for their training or evaluation; (2) they are fast and efficient; and finally; (3) they require low computational resources. However, string-based methods are unable to capture the semantics of the words in the sentence, which prevent them from recognizing semantic relationships, such as synonymy, meronymy and morphological variants. On the other hand, the use of NER tools in combination with string-based methods is a good option to integrate at least the capability of recognizing synonyms, as shown by LiBlocK-CTakes (M4).

\emph{Ontology-based methods strongly depends on the lexical coverage of the ontologies and the ability to recognize automatically the underlying concepts in sentences.} Our results show that the ontology-based methods are able to properly estimate a similarity score when it is evaluated in a dataset with either high word overlapping or NER and WSD tools that find all possible entities to properly calculate the similarity between sentences. The main advantages of ontology-based methods are that they are fast and require low computational resources. However, the effectiveness of the ontology-based methods depends on the lexical coverage of the ontologies and the ability of the NER and WSD tools to recognize the underlying concepts in sentences, whose coverage and performance could be limited in several application domains. 

The LiBlock (M4) string-based method and the COM (M17) ontology-based method use a NER tool in the pre-processing stage to recognize the biomedical entities (UMLS CUI codes) present in the input sentences. The objective of annotating entities in the semantic similarity task is the identification and disambiguation of biomedical concepts to provide semantic information to sentences. LiBlock uses the NER tool to normalize and disambiguate the underlying concepts in a sentence, unifying different concepts with acronyms and synonyms in the same CUI code and creating an overlapping between concepts, while ontologies also make use of the similarity of concepts within ontologies.

\emph{The biomedical NER tools evaluated in this work are unable to identify and disambiguate correctly many biomedical concepts due to the use of acronyms and different morphological variations, among others.} For example, the CUI concepts ``KRAS gene" (C1537502), ``BRAF gene" (C0812241), and ``RAF1 gene" (C0812215) in the sentences $s1$ and $s2$ with highest $E_{sim}$ obtained by the COM (M17) method in table \ref{tab:raw_best_worst_sentences_ont}, appear as ``K-ras", ``Braf", ``c-Raf" and ``Craf'. However, cTakes is unable of recognizing these later morphological variants of the same biomedical concepts. A second example is the word ``act" in the sentence ``Consequently miRNAs have been demonstrated to act either as oncogenes [...]", which is wrongly recognized as the entity ``Activated clotting time measurement" (C0427611), rather than as a verb in the sentence $s1$ with highest $E_{sim}$ in table \ref{tab:raw_best_worst_sentences_string}. And finally, a third example is the acronym ``NSCLC", which denotes the concept ``Non-Small Cell Lung Carcinoma (C0007131), which is not recognized in the plural variant ``NSCLCs" in the sentence $s2$ with highest $E_{sim}$ from table \ref{tab:raw_best_worst_sentences_ont}.

The methods based on pre-trained embeddings and language models provide a broader lexical coverage than the ontology-based methods, and do not need the use of NER or WSD tools to find intrinsic semantic relationships between the words in the sentences. However, these later methods need large corpus for their training, as well as a complex training phase and more computational resources than the methods from the families of string-based and ontology-based. On the other hand, our experiments show that those methods tend to estimate higher similarity values than those estimated by a human being in the three datasets. In most cases, the aforementioned method report similarity scores that tend to 1, which indicates that the semantics obtained from the sentences is not sufficient to compute correctly a similarity score. For instance, the sentences $s1$ and $s2$ with highest $E_{sim}$ from tables \ref{tab:raw_best_worst_sentences_embedd} and \ref{tab:raw_best_worst_sentences_bert} shows similarity values close to 1, where the sentences have neither word overlapping nor similar concepts, and the human similarity score is 0 in both cases. On the other hand, BERT-based methods are trained for downstream tasks, using a supervised approach, and do not perform well in an unsupervised context.

\paragraph{Answering RQ5.} String-based methods capture neither the word semantics within the sentences nor the semantic relationships between words, such as synonymy and meronymy, and their effectiveness mainly relies on the word overlapping frequency in the sentences. However, the LiBlock method uses the NER tool to normalize and disambiguate the underlying concepts in a sentence, but unfortunately, it does not significantly outperform LiBlock with no use of a NER tool, which could be caused by two reasons as follows. Firstly, the incapability of LiBlock to capture semantic relationships beyond the synonymy, and secondly the current limitations of cTakes to recognize all mentions of biomedical entities. On the other hand, ontology-based methods use NER and WSD tools to recognize the underlying concepts in the sentences, which are not able to correctly identify and disambiguate these concepts in many cases. In addition, they require external resources to capture the semantic information from the sentences, which limits their lexical coverage. Thus, ontology-based methods require both high word overlapping and high recognition coverage of named entities to properly estimate the similarity between sentences. On the other hand, the methods based on pre-trained embeddings and language models need large corpus for training, a complex training phase, and considerable computational resources to calculate the similarity between sentences. Moreover, those methods tend to obtain high similarity scores in most cases, which may penalize them in a balanced dataset and in a real environment. Finally, BERT-based methods are trained for downstream tasks, using a supervised approach, and do not perform well in an unsupervised context.

\subsection*{Comparison of running times}

Table \ref{tab:exec_times} details the running time reported by the best-performing methods for each family, as well as the sentences per second that computes each method by average for the three datasets evaluated herein. The experiments were executed in a desktop computer with an AMD Ryzen 7 5800x CPU (16 cores) with 64 Gb RAM and 2TB Gb SSD disk. In all the cases, the running time also comprises the pre-processing time for each method. The string-based method Block Distance (M3) obtain the lowest running times because it does not need complex mechanisms or pre-trained models to calculate the similarity between sentences. On the other hand, the BERT-based methods obtain the worst results mainly due to its pre-processing stage, which uses the WordPiece tokenization method.

\begin{table}[h!]
    \centering
    \caption{This table shows the running times in miliseconds (ms) and the average sentences pairs per second (sent/sec) reported by the best-performing method of each family of methods in the evaluation of the 1339 sentence pairs that conform the three datasets. (*) The LiBlock method reports the running times in both NER and noNER versions showing that the efficiency of the method with no NER tool is much higher, despite the fact that there is no statistically significant difference in the results between both pre-processing configurations.}
    \begin{tabular}{llcc}
        ID & Method & Running time (ms) & Sentence pairs / sec \\
        \hline
        M4 & LiBlock-cTAKES & 56605 & 23,66 \\ 
        M4 & LiBlock-noNER (*) & 508 & 2635,83 \\ 
        M3 & Block distance & 308 & \textbf{4347,4} \\ 
        \hline
        M12 & UBSM-Rada & 32341 & 41,40 \\ 
        M17 & COM & 41558 & 32,22 \\ 
        \hline
        M27 & BioWordVec$_{int}$ & 1211 & 1105,69 \\ 
        M32 & BioSentVec & 54706 & 24,48 \\ 
        \hline 
        M47 & ouBioBERT & 575770 & 2,33 \\ 
        M38 & \makecell[c]{BioBERT Large 1.1 \\ (+ PubMed)} & 3312566 & 0,40 \\ 
    \end{tabular}
    \label{tab:exec_times}
\end{table}

\subsection*{Inconsistent results in the calculation of the statistical significance matrix.} Despite the artificial increase of datasets to calculate the statistical significance of the results, we have identified an inconsistent result with respect to the comparison of the p-values of the LiBlock (M4) and the WBSM-Rada (M7) and UBSM-Rada (M12) methods. Table \ref{tab:table_results} shows that the UBSM-Rada method (M12) has a higher average harmonic score compared to WBSM-Rada (M7). However, by building the artificial datasets, the value of UBSM-Rada (M12) with respect to LiBlock (M4) shows a significant difference, while WBSM-Rada (M7) with respect to LiBlock (M4) shows a non-significant difference. We conjecture that this problem could be solved by increasing the number of datasets created for this task, which would allow to increase the sample size and obtain more consistent results.

\section*{Conclusions and future work}

We have introduced the largest, detailed, and for the first time, reproducible experimental survey on biomedical sentence similarity reported in the literature. Our work also introduces a collection of self-contained and reproducible benchmarks on biomedical sentence similarity based on the same software platform, called HESML-STS, which has been especially developed for this work, being provided as part of the new HESML V2R1 version that will be made publicly available soon. We provide a detailed reproducibility protocol \cite{Lara-Clares2022protocolsIO} and dataset \cite{EPNXTR_2021Dataset} to allow the exact replication of all our experiments, methods, and results. In addition, we introduce a new aggregated string-based sentence similarity method called LiBlock, together with eight variants of the ontology-based methods introduced by Sogancioglu et al. \cite{Sogancioglu2017-rc}, and a new pre-trained word embedding model based on FastText \cite{Bojanowski2017-pb} and trained on the full-text of the articles in the PMC-BioC corpus \cite{Comeau2019-vd}. We also evaluate for the first time the CTR \cite{Lithgow-Serrano2019-si} dataset in a benchmark on biomedical sentence similarity.

The string-based LiBlock (M4) measure sets the new state-of-the-art for the sentence similarity task in the biomedical domain and significantly outperforms all the methods evaluated herein, with the only exception of the COM (M17) and WBSM-Rada (M7) ontology-based methods. However, our data analysis shows that at least with the three datasets evaluated herein, there is no statistically significant difference between the performance of the LiBlock (M4) method using the cTakes or none NER tool. Thus, using the LiBlock method without any NER tool could be a competitive and much more efficient solution for high-throughput applications.

Concerning the impact of the Named Entity Recognition (NER) tools, our results confirm that the choice of the best NER tool for each method significantly impacts their performance. MetamapLite \cite{Demner-Fushman2017-zs} and cTAKES \cite{Savova2010-ed} set the best-performing configurations for the family of ontology-based methods, whilst Metamap \cite{Aronson2010-pb} sets the best-performing option for none.

Our experiments confirm that the pre-processing stage has a very significant impact on the performance of the sentence similarity methods evaluated herein, despite this fact have neither been studied nor reported in the literature. Thus, the selection of the proper configuration for each sentence similarity method should be confirmed experimentally. However, our experiments suggest some default configurations to make these decisions, such as the use of lower-casing normalization, some specific char filtering methods, and some specific tokenizers with the only exception of BioCNLPTokenizer. Finally, the families of string and ontology-based methods show a noticeable preference pattern by the use of the NLTK2018 stop-words list. For a detailed description of the best pre-processing configurations, we refer the readers to our discussion.

String-based methods do not capture either the semantics of the words in the sentence or the semantic relationships between words, and their effectiveness relies on the word overlapping frequency in the sentences. Ontology-based methods Named Entity Recognition (NER) and Word Sense Disambiguation (WSD) tools to recognize the underlying concepts in the sentences and require external resources to capture the semantic information from the sentences, which limits their lexical coverage. In addition, they require either high word overlapping or high recognition coverage of named entities in order to properly calculate the similarity between sentences. On the other hand, the methods based on pre-trained embeddings and language models need a large corpus for training, a complex training phase, and considerable computational resources to calculate the similarity between sentences. Moreover, these methods tend to obtain high similarity scores in most cases, which may penalize them in a balanced dataset and in a real environment. Finally, BERT-based methods are trained for downstream tasks, using a supervised approach, and do not perform well in an unsupervised context.

Our experiments suggest that the current benchmarks do not cover all the language features that characterize the biomedical domain, such as the frequent use of acronyms and rhetorical expressions like synonymy, meronymy, etc. In addition, current benchmarks have a very limited sample size that difficult the analysis of results. We conjecture that LiBlock, COM, and UBSM-Rada perform well because there is a noticeable overlap of terms that may benefit the former methods over the others reported in the literature. Furthermore, Chen et al. \cite{Chen2021-kz} highlights the need to improve and create new benchmarks from different perspectives, to reflect the multifaceted notion of the similarity of sentences. Therefore, we found a strong need for improving existing benchmarks for the task of semantic similarity of sentences in the biomedical domain.

As forthcoming activities, we plan to publish our new software release HESML V2R1 including the HESML-STS software package developed for this work. We also plan to evaluate the new sentence similarity methods introduced herein in a benchmark for the general language domain. In addition, we will study the evaluation of the sentence similarity methods in an extrinsic task, such as semantic medical indexing \cite{Couto2020-ct} or summarization \cite{Mishra2014-pa}. We also consider the evaluation of further pre-processing configurations, such as biomedical NER systems based on recent Deep Learning techniques \cite{Hahn2020-os}, or extending our experiments and research to the multilingual scenario by integrating multilingual biomedical NER systems like Cimind \cite{Cabot2019-dm}. Finally, we plan to evaluate some recent biomedical concept embeddings based on MeSH \cite{Abdeddaim2019-mc}, which has not been evaluated in the sentence similarity task yet.

\section*{Acknowledgments}

We are grateful to Gizem Sogancioglu and Kathrin Blagec for answering kindly our questions to replicate their methods and experiments, Fernando Gonz\'{a}lez and Juan Corrales for setting up our reproducibility dataset, and  Hongfang Liu and Yanshan Wang for providing us the MedSTS dataset. UMLS CUI codes, SNOMED-CT US ontology and MeSH thesaurus were used in our experiments by courtesy of the National Library of Medicine of the United States.

\appendix

\section*{Appendix A. The statistical significance results}
\label{appendixa}

We provide a series of tables reporting the p-values for each pair of methods evaluated in this work as supplementary material.

\section*{Appendix B. The pre-processing raw output files}
\label{appendixb}

We provide all the pre-processing raw output tables for the experiments evaluated herein as supplementary material

\section*{Appendix C. A reproducibility protocol and dataset on the biomedical sentence similarity}
\label{appendixc}

We provide the reproducibility protocol published at protocols.io \cite{Lara-Clares2022protocolsIO} as supplementary material to allow the exact replication of all our experiments, methods, and results.

\nolinenumbers

%
%
%

\bibliography{bibliography}

\end{document}